\RequirePackage{fix-cm}
\documentclass[twocolumn]{svjour3}          % twocolumn
\smartqed  % flush right qed marks, e.g. at end of proof
\usepackage{graphicx}
\emergencystretch=\maxdimen
\usepackage{natbib} % for citation
\usepackage{comment}
\usepackage{nicefrac}       % compact symbols for 1/2, etc.
\usepackage{microtype}      % microtypography

\usepackage{xcolor}
\usepackage{color, colortbl}
\usepackage{url}       % url citation
\usepackage{soul} % delete line
\usepackage{multirow}
\usepackage{graphicx}

\usepackage{caption}
\usepackage{subfigure}
\usepackage{array}
\newcolumntype{x}[1]{>{\centering\arraybackslash\hspace{0pt}}p{#1}}

\usepackage{fancyhdr}%导入fancyhdr包
\usepackage{amsmath}
\usepackage{amsfonts}
\usepackage{amssymb}
\usepackage{bbding}
\usepackage{pifont}
\newcommand{\cmark}{\ding{51}}%
\usepackage[switch]{lineno} 

\newcommand{\xmark}{\ding{55}}% for xmark

\def\argmax{\mathop{\rm arg\,max}\limits}

\newcommand{\myparagraph}[1]{\vspace{2pt}\noindent{\bf #1}}

\newcommand*{\Scale}[2][4]{\scalebox{#1}{$#2$}}%

\newcommand{\add}[1]{\textcolor{black}{#1}}
\newcommand{\new}[1]{\textcolor{black}{#1}}

\newcommand{\revise}[1]{\textcolor{black}{#1}}

\begin{document}
% \linenumbers

\title{Attribute Prototype Network for Any-Shot Learning%\thanks{Grants or other notes
}
% \subtitle{Do you have a subtitle?\\ If so, write it here}


\author{Wenjia Xu$^{1,7,8}$ \and Yongqin Xian$^{3}$ \and Jiuniu Wang$^{6,7,8}$ \and Bernt Schiele$^{2}$  \and Zeynep Akata$^{2,4,5}$
}


\institute{
 \small
Wenjia Xu \\ \email{xuwenjia16@mails.ucas.ac.cn} \\
  \\
Yongqin Xian \\ \email{yongqin.xian@vision.ee.ethz.ch} \\
The majority of the work was done when Yongqin Xian was with Max Planck Institute for Informatics \\
 \\
Jiuniu Wang \\ \email{wangjiuniu16@mails.ucas.ac.cn} \\
 \\
Bernt Schiele \\ \email{schiele@mpi-inf.mpg.de} \\
 \\
Zeynep Akata \\ \email{zeynep.akata@uni-tuebingen.de} \\ 
 \\
$^{1}$ State Key Laboratory of Networking and Switching Technology, Beijing University of Posts and Telecommunications, Beijing, China\\
$^{2}$ Max Planck Institute for Informatics, Saarland Informatics Campus, Saarbr\"ucken, Germany\\
$^{3}$ Computer Vision Laboratory, ETH Zurich, Switzerland \\
$^{4}$ University of T\"ubingen, T\"ubingen, Germany \\
$^{5}$ Max Planck Institute for Intelligent Systems, T\"ubingen, Germany \\
$^{6}$ City University of Hong Kong, Hong Kong\\
$^{7}$ Aerospace Information Research Institute, Chinese Academy of Sciences\\
$^{8}$ School of Electronic, Electrical and Communication Engineering, University of Chinese Academy of Sciences, China\\
}

\date{Received: date / Accepted: date}

\maketitle

\begin{abstract}
\add{Any-shot image classification allows to recognize novel classes with only a few or even zero samples.
% \footnote{This submission extends our paper accepted at Neural Information Processing Systems (NeurIPS) 2020, with the title ``Attribute Prototype Network for Zero-Shot Learning''. We marked all the extensions in blue text in the manuscript. What is new in this submission compared to the conference version is concluded in the cover letter.}
For the task of zero-shot learning, visual attributes have been shown to play an important role, while in the few-shot regime, the effect of attributes is under-explored. To better transfer attribute-based knowledge from seen to unseen classes, we argue that an image representation with integrated attribute localization ability would be beneficial for any-shot, i.e. 
% \bernt{detailed comment: I do not like the `\ie' command - as it highlights some rather unimportant detail in the textflow - I very much prefer `i.e.' - same argument holds for e.g., and others}\wenjia{I have replaced ``\ie'' with ``i.e.'', ``\eg'' with ``e.g.'', and ``\vs'' with ``vs''.} 
zero-shot and few-shot, image classification tasks. }
To this end, we propose a novel representation learning framework that jointly learns discriminative global and local features using only class-level attributes. While a visual-semantic embedding layer learns global features, local features are learned through an attribute prototype network that simultaneously regresses and decorrelates attributes from intermediate features. \new{Furthermore, we introduce a zoom-in module that localizes and crops the informative regions to encourage the network to learn informative features explicitly.} We show that our locality augmented image representations achieve a new state-of-the-art on challenging benchmarks, i.e. CUB, AWA2, and SUN. \revise{As an additional benefit, our model points to the visual evidence of the attributes in an image, confirming the improved attribute localization ability of our image representation. 
The attribute localization is evaluated quantitatively with ground truth part annotations, qualitatively with visualizations, and through well-designed user studies.} 

\end{abstract}

\section{Introduction}
\label{intro}
\add{Visual attributes describe discriminative visual properties of objects shared among different classes. Attributes have been shown to be important for zero- and few-shot learning, i.e. any-shot learning, as they allow semantic knowledge transfer from known classes with abundant training samples to novel classes with only a handful of images.
Most zero-shot learning~(ZSL) methods~\citep{romera2015embarrassingly,CCGS16,ALE,DEM}  rely on image representations extracted from a deep neural network pretrained on ImageNet, and essentially learn a compatibility function between the image representations and attributes. Early solutions for few-shot learning~(FSL) are dominated by metric learning~\citep{snell2017prototypical,vinyals2016matching} which aims to find a better distance function, and meta-learning~\citep{finn2017,finn2018probabilistic,munkhdalai2017meta} which learns better model initialization or optimizer. Focusing on image representations that directly allow attribute localization is relatively unexplored. We argue that any-shot learning can significantly benefit from an image representation that allows localizing visual attributes in images, especially for fine-grained~\citep{tokmakov2019learning,tang2020revisiting} scenarios where local attributes are critical to distinguish two similar categories.
In this work, we refer to the ability of an image representation to localize and associate an image region with a visual attribute as locality. Our goal is to improve the locality of image representations for any-shot learning.} 

While modern deep neural networks~\citep{he2016deep} encode local information and some CNN neurons are linked to object parts~\citep{zhou2018interpreting}, the encoded local information is not necessarily best suited for any-shot learning.
%these object parts are not necessarily discriminative for classification. 
There have been attempts to improve the locality of image representations by learning visual attention~\citep{li2018discriminative,SGMA,tang2020revisiting} or attribute classifiers~\citep{sylvain2019locality}. \citet{SGMA} propose to learn channel-wise attention for bird body parts. Similarly, ~\citet{zhu2019learning} apply the channel grouping model~\citep{zheng2017learning} to learn part-based representations and part prototypes. However, the learned latent attentions/prototypes only localize a small number of object parts, and the semantic meaning of the attentions is inducted by post-hoc observation. 

Although visual attention accurately focuses on some object parts, the discovered parts and attributes are often biased towards training classes due to the learned correlations. 
For instance, the attributes \textit{yellow crown} and \textit{yellow belly} co-occur frequently (e.g. in Yellow Warbler). 
The model may learn such correlations as a shortcut to maximize the likelihood of training data and therefore fail to deal with unseen attributes configurations in novel classes, such as \textit{black crown} and \textit{yellow belly} (e.g. in Scott Oriole), as this attribute combination has not been observed before.
% Such correlations may be learned as a shortcut to maximize the likelihood of training data and therefore fail to deal with unknown configurations of attributes in novel classes such as \textit{black crown} and \textit{yellow belly} (e.g. in Scott Oriole), as this attribute combination has not been observed before. 
%essentially overlooks the negative effect of attribute correlations. 
%This limited locality may lead to a certain bias towards training classes. 

To improve locality and mitigate the above weaknesses of image representations, we develop a weakly supervised representation learning framework that localizes and decorrelates visual attributes. More specifically, we learn local features by injecting losses on intermediate layers of CNNs and enforcing these features to encode visual attributes defining visual characteristics of objects. To achieve this, we learn prototypes in the feature space which define the property for each attribute, 
% \bernt{`in the mean time' does not sound proper english in this context to me. Do you mean `and at the same time'?}
at the same time, the local image features are encouraged to be similar to the corresponding attribute prototype. It is worth noting that we use only class-level attributes and semantic relatedness of them as supervisory signal, in other words, no human-annotated association between the local features and visual attributes is given during training. 
%Once learned, the association will enable the network to localize visual attributes in images.  
% \add{In addition, since semantically related attribute prototypes~(e.g. attributes describing the same body part) should be similar in the feature space, while attributes prototypes with no semantic relation should stay away.} 
We propose to alleviate the impact of incidentally correlated attributes by leveraging their semantic relatedness while learning these local features. 
As an additional benefit, our model points to the visual evidence of the attributes in an image, confirming the improved attribute localization ability of our image representation.
 \revise{The attribute attention map is obtained by measuring the similarity between local image features and attribute prototypes. We evaluate the attribute localization ability quantitatively with the ground truth part annotations, and qualitatively with visualizations. For datasets without attribute location annotations, we propose two user studies to assess the accuracy and semantic consistency of attribute attention maps, and compare our \texttt{APN} model with the baseline visualized by two model explanation methods, \texttt{Grad-CAM} and \texttt{CAM}.}

\add{This paper extends our NeurIPS 2020 conference paper~\citep{xu2020attribute} with the following additional contributions.
(1) We propose to apply the Attribute Prototype Network for the any-shot image classification task, where we improve the locality of image representations. In addition to zero-shot learning, our \texttt{APN} is extended to few-shot learning and the more realistic generalized few-shot learning setting. We evaluate our model under both  N-way-K-shot and all-way scenarios and demonstrate that the local representations encoding semantic visual attributes are beneficial for the any-shot regime in discriminating categories with only a few training samples. (2) In addition to performing classification with the original image, we propose to highlight the informative image features discovered by the attribute prototypes, which helps the network to focus on informative attribute regions and discard the noisy background. (3) We verify the effectiveness of our locality-enhanced image representations on top of five 
% \yongqin{give the exact number}
generative models and demonstrate consistent improvement over the state-of-the-art on three challenging benchmark datasets.
% confirming that the local information encoded in image representation helps to synthesize discriminative features.
\revise{(4) Through qualitative analysis among three benchmark datasets and qualitative ablation study, we demonstrate that our proposed model helps to learn accurate attribute prototypes and produce compact attribute attention maps for each image. Quantitative evaluation indicates that our model outperforms other weakly supervised attribute localization methods by a large margin.}
(5) We propose two well-designed user studies to evaluate the accuracy and semantic consistency of the attribute attention maps, which is an effective evaluation protocol in the absence of ground truth annotations. }

% We show that our model consistently improves the attribute localization ability of the baseline model which does not learn attribute prototypes, by only inspecting the attention maps of attribute prototypes and without using any part annotations during training.

%----------Related work--------------
\section{Related work}
\label{Related}
\myparagraph{Zero-shot learning.} The aim of zero-shot learning is to classify the object classes that are not observed during training~\citep{32_awa}. The key insight is to transfer knowledge learned from seen classes to unseen classes with class embeddings that capture similarities between them. Many classical approaches~\citep{romera2015embarrassingly,CCGS16,ALE,DEM,XASNHS16,wang2017zero} learn a compatibility function between image and class embedding spaces. Recent advances in zero-shot learning mainly focus on learning better visual-semantic embeddings~\citep{liu2018generalized,DEM,jiang2019transferable,cacheux2019modeling,wan2021visual,li2020transferrable} or training generative models to synthesize features~\citep{CLSWGAN,xian2019,ABP,zhu2018generative,kumar2018generalized,schonfeld2019generalized,changpinyo2020classifier}. Those approaches are limited by their image representations, which are often extracted from ImageNet-pretrained CNNs or finetuned CNNs on the target dataset with a cross-entropy loss. 

Despite its importance, image representation learning is relatively under-explored in ZSL. Recently, \citet{ji2018stacked} propose to weigh different local image regions by learning attentions from class embeddings. \citet{SGMA} extend the attention idea to learn multiple channel-wise part attentions.~\citet{sylvain2019locality} show the importance of locality and compositionality of image representations for ZSL. In our work, instead of learning visual attention like \citet{SGMA} and \citet{ji2018stacked}, we propose to improve the locality of image features by learning a prototype network that is able to localize different attributes in an image.

\myparagraph{Few-shot learning.} 
\add{Given a large number of training samples from the base classes, few-shot learning~(FSL) aims to recognize novel classes with a handful of (typically 1–10) labeled examples. The data efficiency problem challenges the traditional classification task, and much effort has been devoted to overcome the data sparsity issue. Metric learning based methods tackle this problem by comparing the distance between images, i.e. Euclidean distance to class prototypes~\citep{snell2017prototypical}, cosine similarity~\citep{vinyals2016matching},
% ridge regression~\citep{bertinetto2018meta}, 
etc. Meta learning based methods aim to learn a good model initialization~\citep{finn2017,finn2018probabilistic,rusu2018meta} or optimizer~\citep{ravi2016optimization,munkhdalai2017meta}, so that the classifiers for novel classes can be learned with a few labeled examples and a small number of gradient update steps. 
% MAML~\citep{finn2017} learns to initialize the model weight so that it can adapt to novel classes efficiently. Meta-LSTM~\citep{ravi2016optimization} learns the optimizer used to train the few-shot classifier. 
However, meta-learning may not sufficiently handle the domain shift between base and novel classes~\citep{chen2019closer}, and has difficulty when scaling to a large number of training samples~\citep{xian2019}. This motivates data augmentation via feature synthesis which directly tackles the data deficiency problem by generating novel training examples~\citep{guan2020zero,xian2019}, e.g. ~\citet{hallucinate2017low} and ~\citet{qi2018low} propose to hallucinate additional training examples for novel classes. }
% ~\citep{imaginarywang2018low} use meta-learning to train the hallucinator directly. 
% In contrast to those prior works that only rely on visual information, 

\add{Existing FSL methods usually rely on prior knowledge from only visual modality, while in zero-shot learning, multi-modality data such as word embeddings~\citep{xian2018zero} and  attributes~\citep{32_awa} have been adopted and achieve promising results. There are some attempts in applying multi-modality to aid representation learning in FSL~\citep{sylvain2019locality,tokmakov2019learning,tang2020revisiting} or to train a feature generator with GANs~\citep{xian2019,guan2020zero}. A few works have focused on enhancing the compositionality~\citep{tokmakov2019learning} with the help of attributes or learning pose-normalized image representations~\citep{tang2020revisiting}, making the representation invariant to the spatial deformations and environmental changes. In this paper, we focus on learning locality-augmented image representations with the help of class attributes to aid the classification task under a low-data regime. We demonstrate that local information encoded in the image representation helps generative models to synthesize discriminative features.}

\myparagraph{Prototype learning.} 
% Our network is motivated by prototypical learning methods~\citep{yang2018robust,mensink2013distance,wang2019panet}.
% Unlike softmax-based CNN, p
Prototype networks~\citep{yang2018robust,wang2019panet} learn a metric space where the labeling is done by calculating the distance between the test image and prototypes of each class.
% similarly as metric based algorithms K-nearest neighbors~(KNN).
Prototype learning is considered to be more robust when handling open-set recognition~\citep{yang2018robust,shu2019p} 
% , out-of-distribution samples~\citep{arik2019attention} 
and few-shot learning~\citep{snell2017prototypical,gao2019hybrid,oreshkin2018tadam}. 
% Prototype learning
Some methods~\citep{arik2019attention,yeh2018representer,li2018deep} base the network classification decision on learned prototypes.
% \yongqin{previous sentence is not precise and is hard to understand.}
% interpret the network decision process~\citep{arik2019attention,yeh2018representer,chen2019looks,li2018deep} in conjunction with several prototypes. Yeh et al.~\citep{yeh2018representer} decompose the network prediction value into a linear combination of prototypes selected from training set, to capture the influence of that training point on the network decision.
% Arik et al.~\citep{arik2019attention} also interpret the network decision based on a few prototype samples from the database. 
Instead of building sample-based prototypes, \citet{chen2019looks} dissect the image and find several prototypical parts for each object category, then classify images by combining evidences from prototypes.
% to improve the model interpretability.
Similarly,~\citet{zhu2019learning} use the channel grouping model~\citep{zheng2017learning} to learn part-based representations and part prototypes.

In contrast, we treat each channel equally and use spatial features associated with input image patches to learn attribute prototypes. \citet{chen2019looks} and \citet{SGMA,zhu2019learning} learn latent attention or prototypes during training and induct the semantic meaning of the prototypes in a post-hoc manner. Their attribute or part localization ability is limited, e.g. \citet{SGMA} can only localize two parts. To address those limitations, our method learns prototypes that represent the attributes/parts where each prototype corresponds to a specific attribute. 
The attribute prototypes are shared among different classes and encourage knowledge transfer from seen classes to unseen classes, yielding better image representation for zero-shot learning and few-shot learning.

\myparagraph{Locality and representation learning.} Here we define the local feature as the image feature encoded from a local image region. Local features have been extensively investigated for representation learning~\citep{hjelm2018learning,wei2019iterative,noroozi2016unsupervised}, and are commonly used in person re-identification~\citep{sun2018beyond,wang2018learning}, image captioning~\citep{updown,li2017image} and fine-grained classification~\citep{zheng2017learning,fu2017look,zhang2016picking}. 
\citet{hjelm2018learning} indicate that maximizing the mutual information between the representation and local regions of the image can significantly improve a representation’s suitability for downstream tasks. 
Thanks to its locality-aware architecture, CNNs~\citep{he2016deep} exploit local information intrinsically. 
% Recent works~\citep{zhou2014object,luo2016understanding} indicate that in deep neural networks such as ResNet50~\citep{he2016deep}, although the theoretical receptive field is much larger than the input image, the effective receptive field can still cover a small image patch. This finding provides practical evidence for associating the local image feature with small input image patches that encode information for a local area.
 Our work is related to methods that draw attention to local features~\citep{kim2018bilinear,sun2018beyond}.
% \add{Sun et al.~\citep{sun2018beyond} propose refined part loss to make sure that each column of pedestrian feature extracts information from the corresponding human body parts. }
% \add{Learning attention over small object parts with bounding box supervision is a popular technique in fine-grained classification tasks~\citep{wei2018mask,zhang2014part,lin2015deep}. To alleviate the supervision, } 
\citet{zheng2017learning} generate the attention for discriminative bird parts by clustering spatially-correlated channels. Instead of operating on feature channels, we focus on the spatial configuration of image features and improve the locality of our representation. Besides, our work is supervised by the class attributes, and no part or bounding box annotation is required. 
% learn prototypes that are close to specific local features. 
% Besides, we propose regularization loss to distinguish the prototypes from different parts and extract images feature with more local information. 
% To evaluate the locality of features, Tristan et al.~\citep{sylvain2019locality} report the mutual information between local and global features in different images. We follow the standard setting in body part detection network SPDA-CNN~\citep{spdacnn,SGMA}, reporting the IoU as the localization accuracy.

\new{Our work is also related to the methods that localize the main object in a weakly supervised way and discard the irrelevant background for image classification~\citep{zhang2021multi,wei2017selective}. \citet{wei2017selective} aggregate the activation map from the last CNN layer~(before global average pooling) and select the highest activated location to filter the final representation for the target image. \citet{zhang2021multi} improve the localization accuracy by introducing multi-layer activation maps. While in our network, we aim to search for the informative image regions that contain important attributes for zero- and few-shot learning. Unlike previous pioneers utilizing intermediate activation maps to localize the informative area, we select the image regions highlighted by the attribute prototypes.} \revise{Since most of the datasets lack ground truth attribute location annotation, e.g. AWA2 and SUN, we design two user studies to assess the accuracy and semantic consistency.}

%----------Methodology--------------
\section{Attribute Prototype Network}
\begin{figure*}[tb]
\label{APN}
  \centering
  \includegraphics[width=\linewidth]{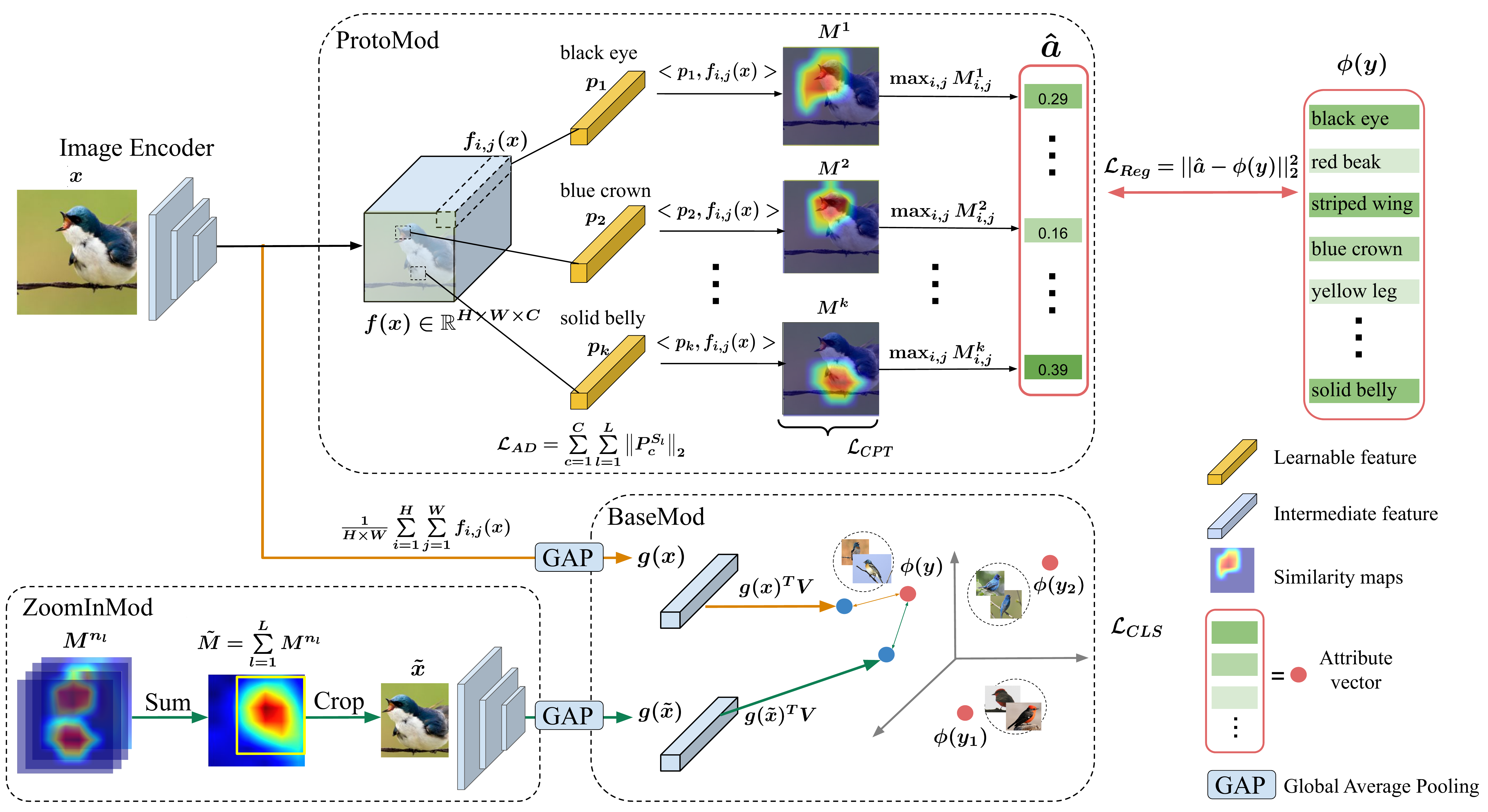}
    \caption{
        \new{Our attribute prototype network~(\texttt{APN}) consists of an \texttt{Image Encoder} extracting image features $f(x)$, a \texttt{BaseMod} performing classification, a \texttt{ProtoMod} learning attribute prototypes $p_k$ and localizing them with similarity maps $M^k$, and a \texttt{ZoomInMod} cropping the informative image region $\tilde{x}$ covered by attribute similarity maps. The end-to-end training of \texttt{APN} encourages the image feature to contain both global information which is discriminative for classification and local information which is crucial to localize and predict attributes.}
    }
  \label{fig:APN}
\end{figure*}

In the following, we describe our end-to-end trained attribute prototype network~(\texttt{APN}) that improves the attribute localization ability of the image representation, i.e. locality. We first define our zero-shot learning and few-shot learning problem. We then introduce in detail the three modules in our framework, the base module~(\texttt{BaseMod}), the prototype module~(\texttt{ProtoMod}), and the Zoom-In Module~(\texttt{ZoomInMod}) as shown in Figure~\ref{fig:APN}. 
At the end of the section, we describe how we perform ZSL and FSL 
% on top of our image representations 
and how the locality enables attribute localization.

\myparagraph{Problem Definition}
The training set consists of labeled images and attributes from seen classes, i.e. $S=\{x, y, \phi(y)| x \in \mathcal{X}, y \in \mathcal{Y}^s \}$.  Here, $x$ denotes an image in the RGB image space $\mathcal{X}$, $y$ is its class label, and $\phi(y) \in \mathbb{R}^{K}$ is the class embedding (i.e. a class-level attribute vector  
annotated with $K$ different visual attributes). 
% \jimmy{``i.e.,'' or ``i.e.''? Please choose one for the whole paper.}
%\add{In addition, we are also given the class embeddings of $\mathcal{Y}^n$, which are the unseen classes~(in zero-shot learning literature)/novel classes~(in few-shot learning literature). }
Here we use $\mathcal{Y}^n$ to denote the unseen class label set in ZSL and the novel class in FSL for convenience. The class embeddings of unseen classes, i.e. $\{\phi(y)| y \in \mathcal{Y}^n \}$, are also known. 
The goal for ZSL is to predict the label of images from unseen classes, i.e. $\mathcal{X} \rightarrow \mathcal{Y}^n$, while for generalized ZSL~(GZSL)~\citep{xian2018zero} the goal is to predict images from both seen and unseen classes, i.e. $\mathcal{X} \rightarrow \mathcal{Y}^n \cup \mathcal{Y}^s$. 
\add{Few-shot learning~(FSL) and generalized few-shot learning~(GFSL) are defined similarly. The main difference lies that instead of only knowing the attributes of novel classes in ZSL, FSL also gets a few training samples from each novel class. }

\subsection{Base Module~(\texttt{BaseMod}) for global feature learning}

%The framework of \texttt{APN} is illustrated in Figure~\ref{fig:APN}. It consists of three module, the \texttt{Image Encoder}, the \texttt{BaseMod} and the \texttt{ProtoMod}. \texttt{Image Encoder} extracts image features from input $x$. \texttt{ProtoMod} learns attribute prototypes and encourage the image features to encode more local information from small image patches. \texttt{BaseMod} learns a compatibility function to maps the image feature into the attribute space and assign each feature to the closest class embedding. 

%As shown in Figure~\ref{fig:APN}, the input image $x$ first passes through the encoder and becomes a $3D$ image feature map $F(x)\in \mathbb{R}^{H\times W \times C}$, where $H$, $W$ and $C$ represents the height, width and channel respectively.  we apply global average pooling to $f(x)$. To learn discriminative global features for ZSL task, we employ softmax classification loss $\mathcal{L}_{CLS}$ to encourage a high inter-class distinction in \texttt{BaseMod}. We apply global average pooling to $F(x)$ to get a single vector
 %$g(x) \in \mathbb{R}^C$:
 
The base module~(\texttt{BaseMod}) learns discriminative visual features for classification. Given an input image $x$, the \texttt{Image Encoder}~(a CNN backbone) converts it into a feature representation $f(x)\in \mathbb{R}^{H\times W \times C}$ where $H$, $W$ and $C$ denote the height, width, and channel respectively.
\texttt{BaseMod} then applies global average pooling over the $H$ and $W$ to learn a global discriminative feature $g(x)\in \mathbb{R}^C$:
\begin{linenomath*}
\begin{equation}
\label{eq:GAP}
    g(x) = \frac{1}{H \times W} \sum_{i=1}^{H} \sum_{j=1}^{W} {f_{i,j}(x)} \,,
\end{equation}
\end{linenomath*}
where $f_{i,j}(x) \in \mathbb{R}^C$ is extracted from the  feature $f(x)$ at spatial location $(i, j)$~(blue box in Figure~\ref{fig:APN}).

% In contrast of standard CNN classifier that uses fully connected layers to compute class logits, we further improve the expressiveness of the image representation by using the attributes values for every seen classes. The annotated attributes values for seen classes are denoted as $\Phi \in \mathbb{R}^{|\mathcal{Y}^s| \times K}$, here $|\mathcal{Y}^s|$ is the number of seen classes. In detail, a linear layer with parameter $V \in \mathbb{R}^{K\times C}$ maps the visual feature $g(x)$ into the semantic embedding~(e.g. attribute) space, as shown in Figure~\ref{fig:APN}. Then the projected visual feature is then multiplied with the attribute matrix $\Phi$ to produce class logits $\hat p$ as
% \begin{equation}
%     \hat p = softmax(\Phi V g(x))\\, 
% \end{equation}
% We apply standard cross-entropy loss after this layer to encourage the image to have the highest compatibility score with its corresponding attribute embedding as, 
% \begin{equation}
%     \mathcal{L}_{CLS}  = -\log \hat p_{gt}\\,
% \end{equation}
% where $\hat p_{gt}$ is the value corresponding to the ground truth class in $\hat p$. The visual-semantic embedding layer and the CNN backbone are optimized jointly to finetune the image representation guided by the attribute embeddings. 

\textbf{Visual-semantic embedding layer.} In contrast to standard CNNs with fully connected layers to compute class logits, we further improve the expressiveness of the image representation using attributes. In detail, a linear layer with parameter $V \in \mathbb{R}^{C\times K}$ maps the visual feature $g(x)$ into the class embedding~(e.g. attribute) space. 
The dot product between the projected visual feature and every class embedding is computed to produce class logits $s = g(x)^T V \phi(y)$, followed by the cross-entropy loss encouraging the image to have a high compatibility score with its corresponding attribute vector.
%The compatibility between projected visual feature~(blue dot in Figure~\ref{fig:APN}) and the ground truth attribute vectors of seen classes is maximized using the standard cross-entropy loss to encourage the image to have the highest compatibility score with its corresponding attribute vector. 
Given a training image $x$ with a label $y$ and an attribute vector $\phi(y)$, the classification loss $\mathcal{L}_{\Scale[0.6]{CLS}}$ is:
%
% \begin{equation}
%     \mathcal{L}_{\Scale[0.6]{CLS}}  = -\log \frac{\exp(g(x)^T V \phi(y))}{\sum_{\hat{y} \in \mathcal{Y}^s}\exp(g(x)^T V \phi(\hat{y}))} \,.
% \end{equation}
\begin{linenomath*}
    \begin{equation}
        \mathcal{L}_{\Scale[0.6]{CLS}}  = -\log \frac{\exp(s)}{\sum_{\mathcal{Y}^s}\exp(s_j)} \,,
        \label{equ:CE}
    \end{equation}
\end{linenomath*}
where $s_j = g(x)^T V \phi(y_j)$, $y_j \in \mathcal{Y}^s$.
The visual-semantic embedding layer and the CNN backbone are optimized jointly to finetune the image representation guided by the attribute vectors. 

%\suggest{Suggestions from Stephan: I only understood this sentence after looking at the formula. It is fine, but it misses a bit the fact that after projecting the image features into the semantic embedding space with V, you take the dot product with the ground truth embedding vector for the class to product a scalar "compatibility score" (logit). On this score you apply the cross entropy loss. I was confused because this sentence suggests that you apply cross entropy on the embedding space which didn't make sense to me.}

\subsection{Prototype Module~(\texttt{ProtoMod}) for local feature learning}

The global features learned from \texttt{BaseMod} may be biased to seen classes because they mainly capture global context, shapes and other discriminative features that may be indicative of training classes. To improve the locality of the image representation, we propose a prototype module~(\texttt{ProtoMod}) focusing on the local features that are often shared across seen and unseen classes.
%that enables local features to localize visual attributes shared across seen and unseen classes. 

\begin{comment}
\myparagraph{Attribute prototypes.} \texttt{ProtoMod} takes as input $f(x)\in \mathbb{R}^{H\times W \times C}$ coming from the \texttt{Image Encoder} composed by the local features $f_{i,j}(x)\in \mathbb{R}^{C}$ that encode the information of local image regions where local attributes and object parts appear.
We learn a set of attribute prototypes $P = \left \{ p_k \in \mathbb{R}^{C} \right \}_{k=1}^K$ to predict attributes from those local features, where $p_k$ denotes the prototype for the $k$-th attribute.
% and $K$ is the total number of attributes. 
As a schematic illustration, $p_1$ and $p_2$ in Figure~\ref{fig:APN} correspond to the prototypes for \textit{black eye} and \textit{blue crown} respectively. For each attribute~(e.g. $k$-th attribute), we compute a similarity map $M^k \in \mathbb{R}^{H\times W}$ that measures the compatibility between its prototype $p_k$ and all the local features, where each element of $M^k$ is computed by $M_{i, j}^k=\langle p_k, f_{i,j}(x)\rangle$. Afterwards, we predict the $k$-th attribute $\hat{a}_k$ by taking the maximum value in the similarity map $M^k$:
\end{comment}

\myparagraph{Attribute prototypes.} \texttt{ProtoMod} takes as input the feature $f(x)\in \mathbb{R}^{H\times W \times C}$ produced by the \texttt{Image Encoder} where the local feature $f_{i,j}(x)\in \mathbb{R}^{C}$ at spatial location $(i, j)$ encodes information of local image regions. Our main idea is to improve the locality of the image representation by enforcing those local features to encode visual attributes that are critical for ZSL.
Specifically, we learn a set of attribute prototypes $P = \left \{ p_k \in \mathbb{R}^{C} \right \}_{k=1}^K$ to predict attributes from those local features, where $p_k$ denotes the prototype for the $k$-th attribute. As a schematic illustration, $p_1$ and $p_2$ in Figure~\ref{fig:APN} correspond to the prototypes for \textit{black eye} and \textit{blue crown} respectively. For each attribute~(e.g. $k$-th attribute), we produce a similarity map $M^k \in \mathbb{R}^{H\times W}$ where each element is computed by a dot product between the attribute prototype $p_k$ and each local feature, i.e. $M_{i, j}^k=\langle p_k, f_{i,j}(x)\rangle$. 
%Intuitively, $M^k$ measures the compatibility between its prototype $p_k$ and local features at all locations. %Each element of $M^k$ is computed by a dot product between the attribute prototype $p_k$ and each local feature, $M_{i, j}^k=\langle p_k, f_{i,j}(x)\rangle$.
Afterwards, we predict the $k$-th attribute $\hat{a}_k$ by taking the maximum value in the similarity map $M^k$:
\begin{linenomath*}
    \begin{equation}
        \label{eq:pred_attr}
        \hat{a}_{k} = \max_{i, j} M_{i, j}^k \,.
    \end{equation}
\end{linenomath*}
Other alternative operations such as average pooling and weighted average pooling are also performed for consideration, while max-pooling works the best, since it associates each visual attribute with its closest local feature and allows the network to efficiently localize attributes.
% Therefore, the predicted attribute value is the highest compatibility score of $p_k$ and the local features.
%The highest compatibility score between $p_k$ and the local features is our predicted attribute value.

\myparagraph{Attribute regression loss.} The class-level attribute vectors supervise the learning of attribute prototypes. We consider the attribute prediction task as a regression problem and minimize the Mean Square Error~(MSE) between the ground truth attributes $\phi(y)$ and the predicted attributes $\hat{a}$: 
\begin{linenomath*}
    \begin{equation}
    \label{eq:reg}
        \mathcal{L}_{\Scale[0.6]{Reg}} = ||\hat{a} - \phi(y) ||^2_2 \,,
    \end{equation}
\end{linenomath*}
where $y$ is the ground truth class. By optimizing the regression loss, we enforce the local features to encode semantic attributes, improving the locality of the image representation. 

\myparagraph{Attribute decorrelation loss.} 
Visual attributes are often correlated with each other as they frequently co-occur, e.g. \textit{blue crown} and \textit{blue back} for Blue Jay birds. 
% As a consequence, the attribute prototypes may use these correlations as a shortcut to maximize the likelihood of training data and fail to deal with unknown configurations of attributes in novel classes. 
%Consequently, the network may not be able to tell the attributes apart and fail to recognize unknown combinations of attributes in novel classes.
Consequently, the network may use those correlations as a useful signal and fails to recognize unknown combinations of attributes in novel classes.
Therefore, we propose to constrain the attribute prototypes by encouraging feature competition among unrelated attributes and feature sharing among related attributes. To represent the semantic relation of attributes, we divide all $K$ attributes into $L$ disjoint groups, encoded as $L$ sets of attribute indices $S_1,\dots, S_L$. We directly adopt the disjoint attribute groups defined by the datasets~\citep{26_wah2011caltech,32_awa,25_SUNdataset}.
Two attributes are in the same group if they have some semantic tie, e.g. \textit{blue eye} and \textit{black eye} are in the same group as they describe the same body part, while \textit{blue back} belongs to another group.  
%\wenjiaadd{\textit{Blue crown} and \textit{black eye} are in same group since they share close local image patch, while \textit{blue back} belongs to another group.} 
For each attribute group $S_l$, its attribute prototypes $\{p_k|k \in S_l\}$ can be concatenated into a matrix $P^{S_l} \in \mathbb{R}^{C \times |S_l|}$, and $P^{S_l}_c$ is the $c$-th row of $P^{S_l}$. We adopt the attribute decorrelation~(AD) loss inspired from~\citet{jayaraman2014decorrelating}: 
\begin{linenomath*}
\begin{equation}
    \label{eq:ad}
     \mathcal{L}_{\Scale[0.6]{AD}}  =  \sum\limits_{c = 1}^{C} \sum\limits_{l = 1}^{L} {\left \| P^{S_l}_c  \right \|_2} \,.
\end{equation}
\end{linenomath*}
%\wenjiaadd{The $\ell_{2}$ norm $\|P_c^{S_l}\|_2$ allows the channel wise feature sharing of prototypes within $S_l$. Afterwards, $\ell_{1}$ norm is applied on each channel as $\sum\nolimits_{c = 1}^{C} {\left \| P^{S_l}_c  \right \|_2}$, and each group as $\sum\nolimits_{l = 1}^{L} {\left \| P^{S_l}_c  \right \|_2}$. The sparse property of $\ell_{1}$ norm could guide these prototypes to ignore features from useless channels for this attribute group, focusing more on effective channels.}
This regularizer enforces feature competition across attribute prototypes from different groups and feature sharing across prototypes within the same groups, which helps decorrelate unrelated attributes.

\myparagraph{Similarity map compactness regularizer.} 
In addition, we would like to constrain the similarity map such that it concentrates on its peak region rather than disperses on other locations.  Therefore, we apply the following compactness regularizer~\citep{zheng2017learning} on each similarity map $M^k$,
\begin{linenomath*}
\begin{equation}
%  \mathcal{L}_{CPT}  = \sum_{k=1}^{K} \sum_{i=1}^{H} \sum_{j=1}^{W}  {M_{i,j}^k \left [ {\left \| i - \tilde{i} \right \|}^2 + {\left \| j - \tilde{j} \right \|}^2  \right ]} \\,
  \mathcal{L}_{\Scale[0.6]{CPT}}  = \frac{1}{KHW} \sum_{k=1}^{K} \sum_{i=1}^{H} \sum_{j=1}^{W}  {M_{i,j}^k \left [ {\left ( i - \tilde{i} \right )}^2 + {\left ( j - \tilde{j} \right )}^2  \right ]} \,,
\end{equation}
\end{linenomath*}
where $(\tilde{i}, \tilde{j}) = \arg\max_{i, j} M^k_{i, j}$ denotes the coordinate for the maximum value in $M^k$. This objective enforces the attribute prototype to resemble only a small number of local features, resulting in a compact similarity map.

% \myparagraph{Joint global and local feature learning.} Our full model optimizes the CNN backbone, \texttt{BaseMod} and \texttt{ProtoMod} simultaneously with the following objective function, 
% \begin{equation}
% \label{equ:APN_loss}
%     \mathcal{L}_{\Scale[0.6]{APN}}  = \mathcal{L}_{\Scale[0.6]{CLS}} + \lambda_1\mathcal{L}_{\Scale[0.6]{Reg}} +  \lambda_2\mathcal{L}_{\Scale[0.6]{AD}} +  \lambda_3\mathcal{L}_{\Scale[0.6]{CPT}} \,,
% \end{equation}
% where $\lambda_1, \lambda_2$, and $\lambda_3$ are hyper-parameters. The joint training improves the locality of the image representation that is critical for any-shot generalization as well as the discriminability of the features. In the following, we will explain how we perform any-shot inference and attribute localization. 

%The joint training will add soft constraint to the encoder and encourage locality in image representation, thus the image feature will deal better with unfamiliar configurations of attributes in novel setting.

\subsection{Zoom-In Module (ZoomInMod) for attribute prototype-informed feature learning}
\new{Previous works have shown that the informative attributes are critical to the knowledge transfer in zero-shot learning~\citep{liu2019attribute,guo2018zero,liu2014automatic}.
We propose a Zoom-In Module~(\texttt{ZoomInMod}) to highlight the image regions covered by the informative attribute similarity maps and discard the irrelevant image regions.  
% Thanks to the prototype module, now we obtain the attribute attentioned area for zero-shot learning, e.g. the attribute similarity maps. 
Instead of performing classification in \texttt{BaseMod} with the original image $x$~(the orange pipeline in Figure~\ref{fig:APN}), the Zoom-In Module crops out the illuminating image region $\tilde{x}$ that are attended by informative attributes and feed the image into \texttt{BaseMod}~(the green pipeline). As illustrated in Figure~\ref{fig:APN}~(left), we sum up the attribute similarity maps for the most informative attribute in each attribute group to form the attention map $\tilde{M}$:
\begin{linenomath*}
\begin{equation}
    \tilde{M} = \sum_{l=1}^{L} M^{n_l} \,, 
\text{ where  } 
    n_l = \argmax_{k \in S_l} a_k \,.
\end{equation}
\end{linenomath*}
$M^{n_l}$ indicates the attribute similarity map, and $n_l$ is the index of the highest predicted~(most informative) attribute in the $l$-th attribute group (e.g. the attention maps for ``white belly'' in ``belly'' attribute group).} 
% \yongqin{The similarity map M is not entirely clear to me. Is the cropping region or M dynamically changing throughout the training since we also optimize the CNN backbone? If yes, you will have to compute the second derivatives as the cropped input image is dependent on the CNN parameters?} \wenjia{Yes, the cropping region and M dynamically changing throughout the training.}
\new{We follow ~\citet{zhang2021multi} to binarize the informative attention map $\tilde{M}$ with the average attention value to form a mask $A$:
\begin{linenomath*}
\begin{equation}
A_{i,j} = 
\begin{cases}
1  & \text{ if } \tilde{M}_{i,j} \geq \overline{m} \\
0 & \text{ if } \tilde{M}_{i,j} < \overline{m}
\end{cases} \,, 
\text{ where  } \overline{m} = \frac{1}{HW} \sum_{i=1}^{H}  \sum_{j=1}^{W} \tilde{M}_{i,j} \,.
\end{equation}
\end{linenomath*}
We upsample the binary mask $A$ to the size of the input image, and use the smallest bounding box covering the non-zero area to crop the original image. Then we feed the cropped image $\tilde{x}$ into the \texttt{Image Encoder}. 
% \yongqin{how is the smallest bounding box computed?} 
Note that there are no parameters in the \texttt{ZoomInMod}. When \texttt{ZoomInMod} is working, the \texttt{BaseMod} takes into two inputs, i.e. the original image $x$ and the Zoom-In image $\tilde{x}$, and maps the visual feature $g(x)$ and $g(\tilde{x})$ into the class embedding space with visual-semantic embedding layer $V$. We sum up the class logits for each image to induct the predicted class. So the overall compatibility scores are as follows:
\begin{linenomath*}
\begin{equation}
    s = g(x)^T V \phi(y) + g(\tilde{x})^T V \phi(y) \,,
\end{equation}
\end{linenomath*}
used to optimize the classification loss in Equation~(\ref{equ:CE}).}

\myparagraph{Joint global and local feature learning.} 
Our full model optimizes the CNN backbone, \texttt{BaseMod} and \texttt{ProtoMod} simultaneously with the following objective function,
\begin{linenomath*}
\begin{equation}
\label{equ:APN_loss}
    \mathcal{L}_{\Scale[0.6]{APN}}  = \mathcal{L}_{\Scale[0.6]{CLS}} + \lambda_1\mathcal{L}_{\Scale[0.6]{Reg}} +  \lambda_2\mathcal{L}_{\Scale[0.6]{AD}} +  \lambda_3\mathcal{L}_{\Scale[0.6]{CPT}} \,,
\end{equation}
\end{linenomath*}
where $\lambda_1, \lambda_2$, and $\lambda_3$ are hyper-parameters. The joint training improves the locality of the image representation that is critical for any-shot generalization as well as the discriminability of the features. In the following, we will explain how we perform any-shot inference and attribute localization. 

\subsection{Zero- and few-shot learning} 

\new{Once our full model is trained, the visual-semantic embedding layer of the \texttt{BaseMod} can be directly used for zero-shot learning inference, which is similar to ALE~\citep{ALE}.
For ZSL, given an image $x$, we generate the ZoomIn image $\tilde{x}$ through the \texttt{ProtoMod} and \texttt{ZoomInMod}, and feed them into the \texttt{BaseMod}. The classifier searches for the class embedding with the highest compatibility via
\begin{linenomath*}
\begin{equation}
    \hat{y} = \argmax_{\tilde{y} \in \mathcal{Y}^n} ~ (g(x)^\mathrm{T}V\phi(\tilde{y}) + g(\tilde{x})^\mathrm{T}V\phi(\tilde{y}))\,.
\end{equation}
\end{linenomath*}
For generalized zero-shot learning~(GZSL), we need to predict both seen and unseen classes. The extreme data imbalance issue will result in predictions to be biased towards seen classes~\citep{chao2016empirical}. To fix this issue, we apply Calibrated Stacking~(CS)~\citep{chao2016empirical} to reduce the seen class scores by a constant factor.
Specifically, the GZSL classifier is defined as,
\begin{linenomath*}
\begin{equation}
    \hat{y} = \argmax_{\tilde{y} \in \mathcal{Y}^n \cup \mathcal{Y}^s} ~ (g(x)^\mathrm{T}V\phi(\tilde{y}) + g(\tilde{x})^\mathrm{T}V\phi(\tilde{y})) - \gamma \mathbb{I} [\tilde{y} \in \mathcal{Y}^s] \,,
\end{equation}
\end{linenomath*}
where the indicator $\mathbb{I}[\cdot]=1$ if $\tilde{y}$ is a seen class and 0 otherwise,  $\gamma$ is the calibration factor tuned on a held-out validation set.}
% For generalized zero-shot learning, the prediction will be biased towards seen classes due to data imbalance issue~\citep{chao2016empirical}. We apply the Calibrated Stacking~(CS)~\citep{chao2016empirical} to reduce the seen class scores by a constant factor $\gamma$.

Our model aims to improve the image representation for novel class generalization and is applicable to other ZSL methods~\citep{ABP,CLSWGAN,CCGS16}, i.e. once learned, our features can be applied to any ZSL model~\citep{ABP,CLSWGAN,CCGS16}. Therefore, in addition to the above classifiers,  we use image features $g(x)$ extracted from the \texttt{Image Encoder}, and train several state-of-the-art ZSL approaches on top of our features, e.g. ABP~\citep{ABP}, f-VAEGAN-D2~\citep{xian2019}, and TF-VAEGAN~\citep{tfvaegan}.

\add{Our APN network can be adapted to the task of few-shot learning~(FSL) by replacing the feature extraction network in FSL methods with our APN network. During the representation learning stage, we train the feature extractor $f(\cdot)$ using the training examples in the base classes $S=\{x, y, \phi(y)| x \in \mathcal{X}, y \in \mathcal{Y}^s \}$, and train the network with the original FSL training loss as well as our $\mathcal{L}_{\Scale[0.6]{APN}}$ loss.}
% \begin{equation}
% \label{equ:APN_loss}
%     \mathcal{L}_{\Scale[0.6]{FSL}} = \mathcal{L}_{\Scale[0.6]{APN}} + \mathcal{L}_{\Scale[0.6]{train}}\\.
% \end{equation}
\add{With the help of the attribute prototype network, we can learn locality augmented representations that are discriminative for FSL models~\citep{yang2020dpgn} and boost their performance.}
\add{Besides, the locality augmented image representations in our model are applicable to the data synthesis based few-shot learning methods~\citep{xian2019, guan2020zero}. We use image features $g(x)$ extracted from the \texttt{Image Encoder} to train several state-of-the-art generative FSL approaches~\citep{tfvaegan,xian2019} and improve their performance.}

\subsection{Attribute localization} 
\label{sec:Attri_loc}
As a benefit of the improved local features, our approach is capable of localizing different attributes in the image by inspecting the similarity maps produced by the attribute prototypes. 
% More specifically, by upsampling the similarity map $M^k$ to the size of the input image with bilinear interpolation, the area with the maximum responses will be the image region corresponding to the $k$-th attribute.
More specifically, we upsample the similarity map $M^k$ to the size of the input image with bilinear interpolation. The area with the maximum responses then encodes the image region that gets associated with the $k$-th attribute.
Figure~\ref{fig:APN} illustrates the attribute regions of \textit{black eye}, \textit{blue crown} and \textit{solid belly} from the learned similarity maps. It is worth noting that our model only relies on class-level attributes and semantic relatedness of them, i.e. attribute groups, as the auxiliary information and does not need any annotation of part locations.

%----------Experiments--------------

% \begin{table}[t]
% \centering
% \resizebox{0.9\linewidth}{!}{%
% \begin{tabular}{c|c|c|c|c}
%  Dataset & \# Images & \# Att & \# Groups & \# $\mathcal{Y}^s/\mathcal{Y}^n$  \\ \hline
%  SUN~\citep{25_SUNdataset} & 14,340 & 102 & 4 & 645/72  \\ %\hline
%  CUB~\citep{26_wah2011caltech} & 11,788 & 312 & 7 & 150/50 \\ %\hline
%  AWA2~\citep{32_awa} & 30,475 & 85 & 9 & 40/10  \\ 
% \end{tabular}
% }
% \caption{The statistics of three benchmark datasets.}
% \label{tab:dataset}
% \end{table}

\begin{table*}[t]
\setlength{\tabcolsep}{4pt}
\renewcommand{\arraystretch}{1.2}
\centering
\resizebox{.85\linewidth}{!}{
\begin{tabular}{l|ccc| cccccc | c}
    %\toprule
    & \multicolumn{3}{c|}{\textbf{ZSL}} & \multicolumn{7}{c}{\textbf{Part localization on CUB}} \\
    Method & CUB & AWA2 & SUN & breast & belly & back  &  head  & wing  & leg   & Mean        \\
    \hline
    $\texttt{BaseMod}$   &  $70.0$  & $64.9$ & $60.0$  & $40.3$  & $40.0$ & $27.2$ & $24.2$ & $36.0$ & $16.5$ & $30.7$  \\
    
    + $\mathcal{L}_{\Scale[0.6]{Reg}}$ & $71.5$ & $66.3$  & $60.9$ & $41.6$ & $43.6$  & $25.2$ & $38.8$ & $31.6$    & $30.2$ & $35.2$ \\
    
     \hspace{2mm}+ $\mathcal{L}_{\Scale[0.6]{AD}}$ & $71.8$ & $67.7$  & $61.4$ & $60.4$ & $52.7$  & $25.9$ & $60.2$ & $52.1$    & $42.4$ & $49.0$  \\
    
     \hspace{4mm}+ $\mathcal{L}_{\Scale[0.6]{CPT}}$ &$72.0$  & $68.4$ & $\textbf{61.6}$ &$63.1$  & $54.6$ & $\textbf{30.5}$ & $64.1$ & $\textbf{55.9}$ & $\textbf{50.5}$ & $52.8$ \\
     
      \hspace{6mm}+ $\texttt{ZoomInMod}$ &$\textbf{75.0}$  & $\textbf{69.9}$ & $61.5$ &$\textbf{67.8}$  & $\textbf{55.9}$ & $29.4$ & $\textbf{68.7}$ & $49.2$ & $49.1$ & $\textbf{53.4}$
     
    %\bottomrule
    \end{tabular}}
\caption{\new{Ablation study of ZSL on CUB, AWA2, SUN~(left, top-1 accuracy) and part localization on CUB~(right, PCP). We train a single \texttt{BaseMod} with the original input $x$ as the baseline. Note that the last row represents our full model \texttt{APN}, which combines \texttt{BaseMod} and \texttt{ProtoMod}~(trained with $\mathcal{L}_{\Scale[0.6]{CLS}}$, $\mathcal{L}_{\Scale[0.6]{Reg}}$, $\mathcal{L}_{\Scale[0.6]{AD}}$, $\mathcal{L}_{\Scale[0.6]{CPT}}$), and the \texttt{ZoomInMod}. }
% (br: breast, be: belly, ba: back,  he: head, wi: wing, le: leg).
}
\label{tab:ablation}
\end{table*}

\section{Experiments}
\label{Experiments}
In the following, we first introduce the datasets. Then we evaluate our attribute prototype network in both zero-shot learning as well as the attribute localization performance. We finally evaluate our network on few-shot learning.

\myparagraph{Datasets.}
We conduct the experiments on three widely used benchmark datasets. CUB~\citep{26_wah2011caltech} is a fine-grained dataset containing $11,788$ images from 200 bird classes with $312$ attributes. Following \texttt{SPDA-CNN}~\citep{spdacnn}, we define $7$ body parts for all the birds in CUB dataset, i.e. \textit{belly}, \textit{breast}, \textit{back}, \textit{wing}, \textit{head}, \textit{leg}, \textit{tail}.
% As shown in Table~\ref{tab:CUB_group}, 
Part related attributes are divided into 
% \bernt{`those' is a bit strange here - I would drop that word}  
seven part groups.
SUN~\citep{25_SUNdataset} is a fine-grained dataset consisting of $14,340$ images from $717$ scene classes, with $102$ attributes divided into $4$ groups, describing the \textit{functions}, \textit{materials}, \textit{surface properties} and \textit{spatial envelope} of scene images.
AwA2~\citep{xian2018zero} is a coarse-grained dataset containing $37,322$ images of $50$ animal classes with $85$ 
% \jimmy{use the same style for the numbers.} 
attributes. We follow~\citet{32_awa} to divide $85$ attributes into $9$ groups, describing various properties of animals, i.e. \textit{color}, \textit{texture}, \textit{shape}, \textit{body parts}, \textit{behaviour}, \textit{nutrition}, \textit{activity}, \textit{habitat} and \textit{character}. For zero-shot learning, we split the seen/unseen classes following~\citet{xian2018zero} %, and details are shown in Table~\ref{tab:dataset}. 
%The seen/unseen(base) classes split for any-shot learning are shown in Table~\ref{tab:dataset}. 
%Following~\citep{xian2019}, we conduct FSL experiments with the standard ZSL splits, 
to avoid overlap between novel images and ImageNet 1K images.

\new{For few-shot learning, we follow two kinds of evaluation protocols. In the N-way-K-shot scenario, the task is to train a classifier with $K$ samples from $N$ classes, to correctly classify the query samples. Following \citet{yang2020dpgn}, we randomly split the CUB dataset into 100 base, 50 validation, and 50 novel classes. In addition to the widely used N-way-K-Shot benchmarks, we also focus on a more realistic setting, i.e. the all-way benchmark~\citep{tang2020revisiting,imaginarywang2018low,hallucinate2017low}. The classifiers are supposed to recognize all the categories simultaneously, i.e. $\mathcal{X} \rightarrow \mathcal{Y}^n$ for FSL and $\mathcal{X} \rightarrow \mathcal{Y}^n \cup \mathcal{Y}^s$ for generalized FSL. In the all-way scenario, we follow~\citet{xian2019} to split three benchmark datasets, i.e. CUB, AWA2, and SUN.}

\begin{figure*}[t]
  \centering
\includegraphics[width=.9\linewidth]{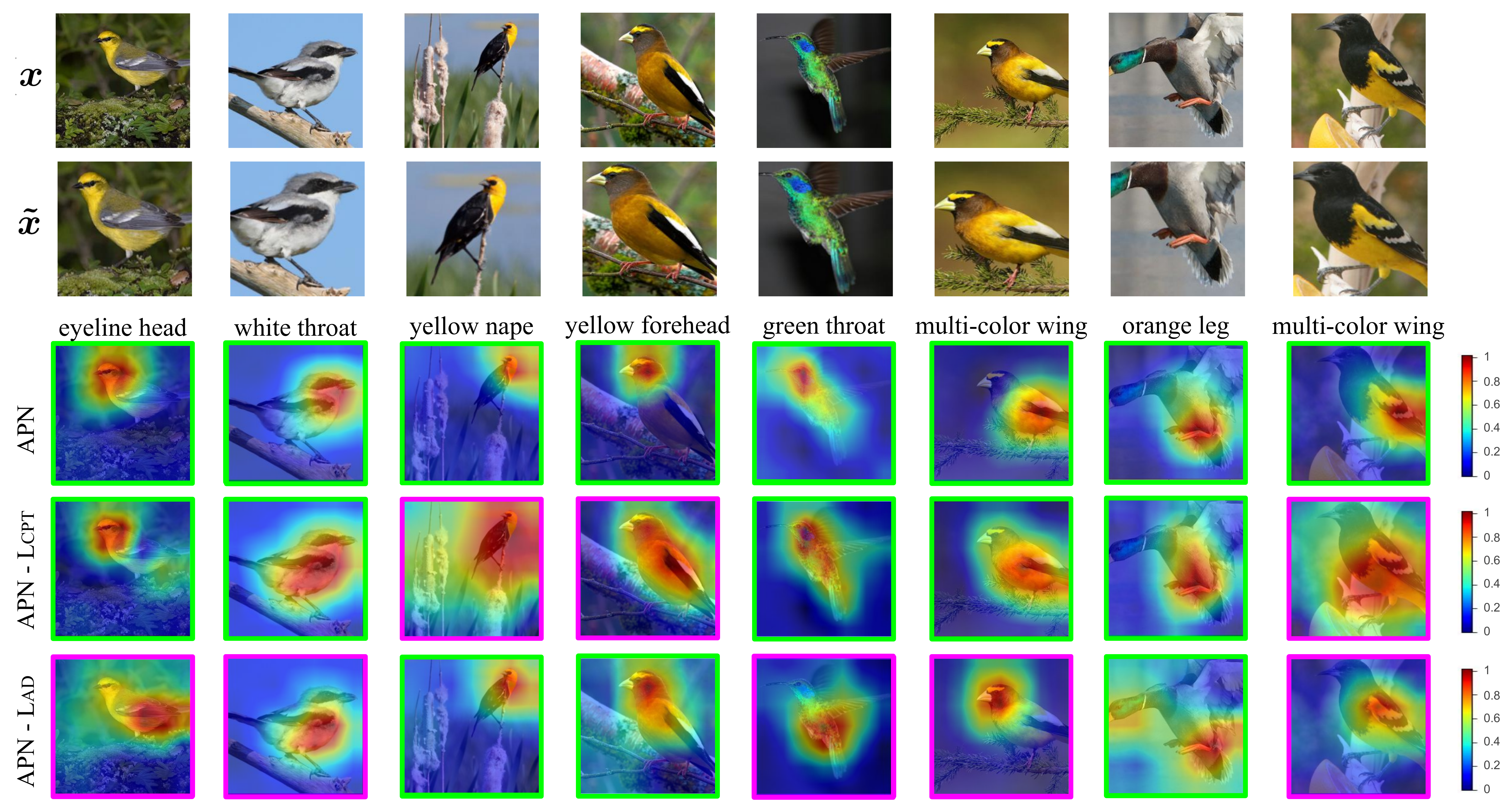}
    \caption{\new{The qualitative ablation study. We display the original image, the Zoom-In image from \texttt{ZoomInMod} in the first and second row. From the third to the fifth row, we show the attribute similarity maps from our \texttt{APN} model, our model trained without $\mathcal{L}_{\Scale[0.6]{CPT}}$, and our model trained without $\mathcal{L}_{\Scale[0.6]{AD}}$, respectively. The text above the attribute similarity maps indicates the attribute name. Green~(purple) box outside the image indicates a correct~(incorrect) localization by our model.}
    }
  \label{fig:ad_cpt_ablation}
\end{figure*}

\setlength{\tabcolsep}{4pt}
\renewcommand{\arraystretch}{1.2} 
\begin{table*}[t]
\centering
\small
\resizebox{\linewidth}{!}
{\begin{tabular}{ll| x{1.2cm} x{1.2cm} x{1.2cm} |c c c |c c c |c c c}
   %\toprule
     & & \multicolumn{3}{c|}{\textbf{Zero-Shot Learning (ZSL)}} & \multicolumn{9}{c}{\textbf{Generalized Zero-Shot Learning (GZSL)}} \\
     & & \textbf{CUB} & \textbf{AWA2} & \textbf{SUN} & \multicolumn{3}{c}{\textbf{CUB}} & \multicolumn{3}{c}{\textbf{AWA2}} & \multicolumn{3}{c}{\textbf{SUN}}  \\
     & Method & \textbf{T1} & \textbf{T1} & \textbf{T1} &  \textbf{u} & \textbf{s} & \textbf{H} & \textbf{u} & \textbf{s} & \textbf{H} & \textbf{u} & \textbf{s} & \textbf{H}\\
    \hline
     \multirow{5}{*}{$\S$}
    & \texttt{SGMA}~\citep{SGMA} & $71.0$ & $68.8$ & $-$ & $36.7$ & $71.3$ & $48.5$ & $37.6$ & $87.1$ & $52.5$ & $-$ & $-$ & $-$  \\
    %  &\texttt{AREN}~\citep{AREN} & $71.8$ & $67.9$ & $60.6$ & $38.9$ & $78.7$ & $52.1$ & $15.6$ & $92.9$ & $26.7$ & $19.0$ & $38.8$  & $25.5$  \\
     &\texttt{AREN}~\citep{AREN} & $71.8$ & $67.9$ & $60.6$ & $63.2$ & $69.0$ & $66.0$ & $54.7$ & $79.1$ & $64.7$ & $40.3$ & $32.3$  & $35.9$  \\
    %   &\texttt{DAZLE}~\citep{DAZLE} & $-$ & $-$ & $-$ & $56.7$ & $59.6$ & $58.1$ & $60.3$ & $75.7$  & $67.1$ & $52.3$ & $24.3$ & $33.2$   \\
     &\texttt{LFGAA+Hybrid}~\citep{liu2019attribute} & $67.6$ & $68.1$ & $61.5$ & $36.2$ & $80.9$ & $50.0$ & $27.0$ & $93.4$ & $41.9$ & $18.5$ & $40.4$  & $25.3$  \\
    %  &\texttt{Composer}~\citep{} & $69.4$ & $\textbf{71.5}$ & $\textbf{62.6}$ & $63.8$ & $56.4$ & $59.9$ & $62.1$ & $77.3$ & $68.8$ & $55.1$ & $22.0$  & $31.4$  \\
    %  &\texttt{DVBE}~\citep{min2020domain} & $-$ & $-$ & $-$ & $64.4$ & $73.2$ & $68.5$  & $62.7$ & $77.5$ & $69.4$ & $44.1$  & $41.6$ & $\textbf{42.8}$ \\
     &\texttt{APN$^{\ddagger}$}~\citep{xu2020attribute} &  $72.0$ & $68.4$ & $\textbf{61.6}$&  $65.3$ & $69.3$ & $67.2$&  $56.5$ & $78.0$ & $65.5$&  $41.9$ & $34.0$ & $\textbf{37.6}$\\
     
      &\texttt{APN}~(Ours) &  $\textbf{75.0}$ & $\textbf{69.9}$ & $61.5$&  $67.4$ & $71.6$ & $\textbf{69.4}$&  $61.9$ & $79.4$ & $\textbf{69.6}$&  $40.2$ & $35.2$ & $37.5$\\
     \hline
     \multirow{11}{*}{$\dagger$}
    % &\texttt{GAZSL}~\citep{zhu2018generative}  &  $55.8$ & $68.2$ & $61.3$&  $23.9$ & $60.6$ &$34.3$& $19.2$&  $86.5$ & $31.4$ &   $21.7$ & $34.5$ & $26.7$ \\
     &\texttt{LisGAN}~\citep{li2019leveraging}  &  $58.8$ & $70.6$ & $61.7$&  $46.5$ & $57.9$ & $51.6$&  $52.6$ & $76.3$ & $62.3$&  $42.9$ & $37.8$ & $40.2$ \\
    &\texttt{CLSWGAN}~\citep{CLSWGAN} &  $57.3$ & $68.2$ & $60.8$&  $43.7$ & $57.7$ & $49.7$&  $57.9$ & $61.4$ & $59.6$&  $42.6$ & $36.6$ & $39.4$ \\
    &\texttt{FREE}~\citep{free2021} &  $-$ & $-$ & $-$ &  $55.7$ & $59.9$ & $57.7$&  $60.4$ & $75.4$ & $67.1$&  $47.4$ & $37.2$ & $41.7$ \\
    \cline{2-14} 
      & \texttt{ABP$^*$}~\citep{ABP} &  $70.7$ & $68.5$ & $62.6$&  $61.6$ & $73.0$ & $66.8$&  $53.7$ & $72.1$ & $61.6$&  $43.3$ & $39.3$ & $41.2$\\
    %  \rowcolor[HTML]{EFEFEF}
    &\texttt{APN+ABP}~(Ours) &  $73.2$ & $\underline{\textbf{73.9}}$ & $63.4$&  $65.5$ & $74.6$ & $ 69.8$&  $57.4$ & $72.3$ & $64.0$&  $46.2$ & $37.8$ & $41.6$\\
    \cline{2-14} 
     & \texttt{CVC$^*$}~\citep{CVC} &  $70.0$ & $64.6$ & $61.0$&  $61.1$ & $74.2$ & $67.0$&  $57.4$ & $83.1$ & $67.9$&  $36.3$ & $42.8$ & $39.3$ \\
      &\texttt{APN+CVC}~(Ours) &  $71.0$ & $71.2$ & $60.6$&  $62.0$ & $74.5$ & $67.7$&  $63.2$ & $81.0$ & $\underline{\textbf{71.0}}$&  $37.9$ & $45.2$ & $41.2$ \\
      \cline{2-14} 
     & \texttt{GDAN$^*$}~\citep{GDAN} &  $-$ & $-$ & $-$&  $65.7$ & $66.7$ & $ 66.2$&  $32.1$ & $67.5$ & $43.5$&  $38.1$ & $89.9$ & $53.4$ \\
      &\texttt{APN+GDAN}~(Ours) &  $-$ & $-$ & $-$&  $67.9$ & $66.7$ & $67.3$&  $35.5$ & $67.5$ & $46.5$&  $41.4$ & $89.9$ & $\underline{\textbf{56.7}}$ \\
      \cline{2-14} 
     & \texttt{f-VAEGAN-D2$^*$}~\citep{xian2019} &  $72.9$ & $70.3$ & $65.6$&  $63.2$ & $75.6$ & $68.9$&  $57.1$ & $76.1$ & $65.2$&  $50.1$ & $37.8$ & $43.1$ \\
      &\texttt{APN+f-VAEGAN-D2$^*$}~(Ours) & $73.9$ & $73.4$ & $65.9$&  $65.5$ & $75.6$ &$70.2$ &  $62.7$ & $68.9$ &$65.7$&  $49.9$ & $39.3$ & $43.8$\\
      \cline{2-14} 
    & \texttt{TF-VAEGAN$^*$}~\citep{tfvaegan} &  $74.3$ & $73.4$ & $65.4$&  $65.5$ & $75.1$ & $70.0$&  $58.3$ & $81.6$ & $68.0$&  $45.3$ & $40.7$ & $42.8$ \\
     &\texttt{APN+TF-VAEGAN}~(Ours) &  $\underline{\textbf{74.7}}$ & $73.5$ & $\underline{\textbf{66.3}}$&  $65.6$ & $76.3$ & $\underline{\textbf{70.6}}$&  $60.9$ & $79.1$ & $68.8$&  $52.6$ & $37.3$ & $43.7$ \\
\end{tabular}
}
\caption{\new{Comparing our \texttt{APN} model with the state-of-the-art on CUB, AWA2 and SUN datasets. $\dagger$ and $\S$ indicate generative and non-generative representation learning methods respectively. \texttt{AREN}~\citep{AREN} and \texttt{APN}~(Ours) uses Calibrated Stacking~\citep{chao2016empirical} for GZSL. We denote the attribute prototype network trained without \texttt{ZoomInMod} in~\citet{xu2020attribute} as \texttt{APN}$^\ddagger$, and represent our full model with \texttt{ZoomInMod} as \texttt{APN} in this paper. For fair comparison, the generative models marked with $^*$~(e.g. \texttt{ABP$^*$}) use \textit{finetuned features} extracted from ResNet101, while the models below, e.g. \texttt{APN+ABP}(Ours), use our \textit{APN features}. We measure top-1 accuracy~(\textbf{T1}) in ZSL, top-1 accuracy on seen/unseen~(\textbf{s/u}) classes and their harmonic mean~(\textbf{H}) in GZSL.} %\zeynep{something is going wrong with the ZSL table. one of the rows may be expecting another column. all columns should have the same width. please check}
% \wenjia{I will fix the $~\dagger$ covered by gray color problem later.}
}
\label{tab:ZSL_acc}
\end{table*}

\myparagraph{Implementation.} 
% \bernt{I would delete the first three words from the first sentence - they do not say anything really}\wenjia{deleted}
To train the attribute prototype network, we adopt ResNet101~\citep{he2016deep} pretrained on ImageNet~\citep{39_Imagenet} as the backbone, and jointly finetune the entire model in an end-to-end fashion to improve the image representation.
We use the Adam optimizer~\citep{kingma2014adam} with $\beta_1=0.5$ and $\beta_2$=0.999. The learning rate is initialized as $10^{-6}$ and decreased every ten epochs by a factor of $0.9$. Hyperparameters in our model are obtained by grid search on the validation set~\citep{xian2018zero}. We set $\lambda_1$ as ranges from $0.01$ to $0.1$ for three datasets, $\lambda_2$ as $0.01$, and $\lambda_3$ as $0.2$.
% \yongqin{$10^{-9}$ is close to zero, please double check if this is correct.} 
The factor $\gamma$ for Calibrated Stacking is set to $0.7$ for CUB, $0.85$ for AWA2, and $0.4$ for SUN.

\myparagraph{Evaluation metrics}
We follow the same evaluation protocol as demonstrated in~\citet{xian2018zero} and~\citet{xian2019}: for ZSL/FSL we report average top-1 accuracy for unseen~(novel) classes; for GZSL we report average top-1 accuracy for both seen~(s) and unseen~(u) classes, as well as their harmonic mean~(H); for GFSL we report average top-1 accuracy for both seen and novel classes.

% \begin{figure*}[t]
%   \centering
% \includegraphics[width=\linewidth]{IJCV/figures/AD_ablation.pdf}
%     \caption{The qualitative ablation for attribute decorrelation loss $\mathcal{L}_{\Scale[0.6]{AD}}$. We retrieve the top-5 closest images for attribute prototype ``blue forehead''~(left) and ``orange leg''~(right), and show the attribute similarity maps that are generated by our \texttt{APN} model~(row 1,2) and our model trained without $\mathcal{L}_{\Scale[0.6]{AD}}$~(row 3,4), i.e. $\texttt{APN}-\mathcal{L}_{\Scale[0.6]{AD}}$. The red and blue bounding boxes in the image represent the ground truth part bounding box and our results, respectively.
%     }
%   \label{fig:AD_ablation}
% \end{figure*}

\subsection{Zero-shot learning}
In this section, we %first introduce the dataset and experiment settings, then 
present an ablation study of our framework in the ZSL setting, and then we present a comparison with the state-of-the-art in ZSL and GZSL settings.

\myparagraph{Ablation study.} To measure the influence of each model component on the extracted image representation, we design an ablation study where we train a single \texttt{BaseMod} with cross-entropy loss as the baseline, and four variants of \texttt{APN} by adding the \texttt{ProtoMod} and the three loss functions, attribute regression loss $\mathcal{L}_{\Scale[0.6]{Reg}}$, attribute decorrelation loss $\mathcal{L}_{\Scale[0.6]{AD}}$, and compactness regularizer $\mathcal{L}_{\Scale[0.6]{CPT}}$ gradually, and finaly we add the \texttt{ZoomInMod} to generate a ZoomIn image as the input for the \texttt{BaseMod}. 
\new{Our results on CUB, AWA2 and SUN presented in Table~\ref{tab:ablation}~(left) demonstrate that the full \texttt{APN} model improve ZSL accuracy over \texttt{BaseMod} consistently, by $5.0\%$~(CUB), $5.0\%$(AWA2), and $1.5\%$~(SUN).  }
% The main accuracy gain comes from the attribute regression loss and attribute decorrelation loss, which adds locality to the image representation.
The attribute regression loss supervises the learning of each attribute prototype and enforces the local image features to contain attribute information, which boosts the performance by $1.5\%$~(CUB), $1.4\%$~(AWA2), and $0.9\%$~(SUN). This indicates that adding locality to the image representation can help the network to learn discriminative features and significantly improve the performance of unseen classes. The attribute decorrelation loss, which suppresses the unwanted attribute co-occurrence and helps to recognize unknown combinations of attributes in novel classes, provides accuracy gains on AWA2~($1.4\%$), CUB~($0.3\%$), and SUN~($0.5\%$). The compactness loss, which constrains the attribute attention maps, does not influence ZSL accuracy much. \new{The \texttt{ZoomInMod} highlights the informative image region and provides significant performance gain for CUB~($3.0\%$) and AWA2~($1.5\%$).}
% It indicates that the network may use some unwanted correlations as useful signal and fails to recognize unknown combinations of attributes in novel classes.

\revise{The accuracy improvement on SUN is not as great as that on CUB and AWA2 for the following reasons. First, APN aims to learn local image features by regressing and decorrelating attributes, which works well for local and visually-grounded attributes. Most attributes in CUB and AWA2 are related to the local visual properties of the birds and animals. However, the attributes of SUN are designed to be global and abstract~\citep{25_SUNdataset}, thus sometimes can not help with learning locality image features. Second, the data distribution of SUN is quite different from CUB and AWA2. The datasets for bird~(CUB) and animal~(AWA2) classification contain main object and discriminative regions. However, dataset SUN is for scene classification, where each element in a scene is crucial in discriminating it from other categories. For instance, the ``glacier'' in Figure 4~(right, column 2) consists of mountains, snow, and sky, and the global feature of the whole scene can lead to good predictions. Emphasizing local features as APN and the ZoomInMod can not help much.}

\new{The qualitative ablation of our \texttt{APN} network is shown in Figure~\ref{fig:ad_cpt_ablation}. We display the Zoom-In image $\tilde{x}$ generated by the \texttt{ZoomInMod} in the second row. The Zoom-In image $\tilde{x}$ accurately crops out the objects, e.g. the birds in the corner~(row 2, column 3), and discards the noisy background. We qualitatively ablate the attribute decorrelation loss $\mathcal{L}_{\Scale[0.6]{AD}}$ and the compactness regularizer $\mathcal{L}_{\Scale[0.6]{CPT}}$ by visualizing the attribute similarity maps generated by different models. }

\revise{Adding the \texttt{ZoomInMod} increases the model complexity by around two times, e.g. increasing FLOPs from 7.9 GMac to 15.7 GMac, increasing memory from 5,216 MiB to 9,052 MiB. However, the training process is still efficient, which will finish in 48 minutes on CUB dataset with a single NVIDIA V100 GPU.}

\new{From Figure~\ref{fig:ad_cpt_ablation}~(row 3-5), we have the following observations. First, the compactness regularizer $\mathcal{L}_{\Scale[0.6]{CPT}}$ helps the model to generate compact attention maps that focus on the most salient attribute region. For instance, our \texttt{APN} model attends to the correct image region when predicting attributes, and the model works well even when localizing small attribute regions, e.g. the ``yellow nape'', ``yellow forehead'' and ``multi-colored wing'' (row 3, column 3,4,6). While the $\texttt{APN}-\mathcal{L}_{\Scale[0.6]{CPT}}$ model produces disperse attention maps and results in inaccurate attribute similarity maps. When predicting attributes ``yellow nape'' and ``yellow forehead''~(row 4, column 3,4), the model spreads attention over the whole bird body. This observation agrees with the quantitative results in Table~\ref{tab:ablation} where $\mathcal{L}_{\Scale[0.6]{CPT}}$ loss improves the part localization score.}

\new{Second, the decorrelation loss $\mathcal{L}_{\Scale[0.6]{AD}}$ helps the model to avoid attribute correlations and results in precise attention maps. For example, the model $\texttt{APN} - \mathcal{L}_{\Scale[0.6]{AD}}$ misunderstands ``white throat'' with  ``white belly''~(row 5, column 2), and ``green throat'' with  ``green belly''(row 5, column 5). The reason might be the model $\texttt{APN} - \mathcal{L}_{\Scale[0.6]{AD}}$ learns similar prototypes for two attributes since they share often co-occurring color properties. While the model $\texttt{APN}$ trained with attribute decorrelation loss forces attribute prototypes from different body parts to be different and correctly localizes the corresponding area. }

\myparagraph{Comparing with the SOTA.} We compare our attribute prototype network~(\texttt{APN}) with two groups of state-of-the-art models: non-generative models, i.e. \texttt{SGMA}~\citep{SGMA}, \texttt{AREN}~\citep{AREN}, \texttt{LFGAA+Hybrid}~\citep{liu2019attribute}; and generative models, i.e. \texttt{LisGAN}~\citep{li2019leveraging}, \texttt{CLSWGAN}~\citep{CLSWGAN}, \new{\texttt{FREE}~\citep{free2021}}, \texttt{ABP}~\citep{ABP}, \new{\texttt{CVC}~\citep{CVC}}, \new{\texttt{GDAN}~\citep{GDAN}}, \texttt{f-VAEGAN-D2}~\citep{xian2019}, and \new{\texttt{TF-VAEGAN}~\citep{tfvaegan}} on ZSL and GZSL settings. 
As shown in Table~\ref{tab:ZSL_acc}, our \texttt{APN} is 
comparable to or better than SOTA non-generative methods in terms of ZSL accuracy. It indicates that our model learns an image representation that generalizes better to unseen classes.
%Generalized ZSL is a more challenging problem since the model suffers from the extreme data imbalance issue. 
In the more challenging generalized ZSL setting, our \texttt{APN} %suffers from biased data due to the absence of unseen classes, thus the results can be further improved via Calibrated Stacking~(CS)~\citep{chao2016empirical} that balances the data distribution. \texttt{APN + CS}
achieves impressive gains over state-of-the-art non-generative models for the harmonic mean~(H): we achieve $69.4\%$  on CUB and $37.5\%$ on SUN. On AWA2, it obtains $69.6\%$, which is even better than other generative models, e.g. \texttt{LisGAN} with $62.3\%$, \texttt{CLSWGAN} with $59.6\%$, and \texttt{FREE} with $67.1\%$.
This shows that our network is able to balance the performance of seen and unseen classes well, since our attribute prototypes enforce local features to encode visual attributes facilitating more effective knowledge transfer. %; 2) the attribute decorrelation loss that alleviates the issue of biasing towards seen classes.

\add{Image features extracted from our model also boosts the performance of generative models that synthesize CNN image features for unseen classes. We choose five SOTA methods \texttt{ABP}~\citep{ABP}, \texttt{GDAN}~\citep{GDAN}, \texttt{CVC}~\citep{CVC}, \texttt{f-VAEGAN-D2}~\citep{xian2019}, and \texttt{TF-VAEGAN}~\citep{tfvaegan} as generative models, and follow the same training and evaluation protocol as stated in these approaches. For fair comparison, we train these models with \textit{finetuned features}~\citep{xian2019} extracted from ResNet101 (denoted with $^*$). We also report the setting where the feature generating models are trained with our \textit{APN feature} $g(x)$~(e.g.~\texttt{APN + TF-VAEGAN}).} %Overall, our APN features augmented with various generative models yield significant gains over non-generative models, i.e. $2\%$ to $6\%$. \zeynep{point to some specific numbers}

\add{In the ZSL setting, we observe the following. First, for five generative models, our \texttt{APN} consistently boosts their performance on three datasets. For instance, for AWA2, we improve the accuracy of \texttt{ABP} from $68.5\%$ to $73.8\%$, the performance of \texttt{CVC} from $64.6\%$ to $71.2\%$. On the fine-grained datasets CUB and SUN, we also gain performance, e.g. the accuracy of  \texttt{ABP} is improved from $70.7\%$ to $73.3\%$ on CUB, from $62.6\%$ to $63.1\%$ on SUN.}
\revise{This indicates that compared to standard finetuning with only cross-entropy loss, our \textit{APN features} provide more local information helping the generative models to synthesize discriminative image features.}
% Second, compared to other generative models, our models achieve the new state of the art and increase the accuracy by a large margin. For instance, \texttt{APN + TF-VAEGAN~\citep{tfvaegan}} outperforms \texttt{LisGAN} by $15.9\%$ on CUB, $2.9\%$ on AWA2 and $4.6\%$ on SUN in ZSL.
\add{In the GZSL setting, the model predicts both seen and unseen images. %Our \texttt{APN feature} also shows great potential. 
We observe that applying our feature to the generative models consistently boosts the harmonic mean, e.g. we improve \texttt{ABP} by $2.7\%$~(CUB) and $2.3\%$~(AWA2). 
% We also boost the performance of \texttt{f-VAEGAN-D2} on three datasets: $0.9\%$~(CUB) and $1.4\%$~(AWA2) in ZSL; and $1.1\%$~(CUB), $0.6\%$~(SUN) in GZSL.
% We also boost the performance of \texttt{CVC}~\citep{CVC} on three datasets: $0.7\%$~(CUB), $3.1\%$~(AWA2) and $1.9\%$~(SUN), etc. 
% With our \texttt{APN} feature, we achieve the new state of the art on three datasets. 
% Specifically, on AWA2 dataset, our \texttt{APN+CVC} obtains a significantly higher harmonic mean~($71.0\%$) than the the previous state of the art model \texttt{TF-VAEGAN} trained on \textit{finetuned features}~($68.0\%$).
Training with our \texttt{APN} feature achieves a more balanced accuracy with much better performance on unseen classes. Compared to \texttt{CVC$^*$} on AWA2, \texttt{APN+CVC} gains $5.8\%$ on unseen while sacrificing only $2.1\%$ on seen. These results demonstrate that our learned locality-enforced image representation lends itself better for knowledge transfer from seen to unseen classes, as the attribute decorrelation loss achieves de-biasing the label prediction. }

\begin{table*}[t]
\centering
\resizebox{0.9\linewidth}{!}{%
\begin{tabular}{l|c|c|cccccc|c}
%\toprule
Method & Parts Annotation & BB size & Breast & Belly & Back & Head  & Wing  & Leg & Mean         \\
\hline
\texttt{SPDA-CNN}~\citep{spdacnn}&  \multirow{1}{*}{\cmark}  & \multirow{1}{*}{$1/4$}  & $67.5$  & $63.2$ & $75.9$  & $90.9$ & $64.8$ & $79.7$ & $73.6$\\
% Part-based R-CNN\citep{zhang2014part}&  \Checkmark & $61.4$ & \multicolumn{5}{c|}{$70.7$} & $66.1$\\
% \hline
% \texttt{Selective search}~\citep{uijlings2013selective}&   &  & $51.8$  & $51.0$ & $56.1$ & $90.8$ & $62.1$ & $66.3$ & $63.0$\\
% \texttt{Edge box}\citep{zitnick2014edge}&  \Checkmark & $1/4$ & $50.1$  & $48.6$ & $35.7$ & $90.5$ & $53.0$ & $66.3$ & $57.3$\\
% \texttt{MCG}\citep{MCG}&  \Checkmark & $1/4$ & $34.4$  & $34.2$& $43.4$  & $90.6$ & $51.7$ & $53.4$ & $51.3$\\

\hline
\texttt{BaseMod}~(uses $\mathcal{L}_{\Scale[0.6]{CLS}}$)     &  \multirow{2}{*}{\xmark}  & \multirow{2}{*}{$1/4$} & $40.3$  & $40.0$ & $27.2$ & $24.2$ & $36.0$ & $16.5$ & $30.7$  \\

% \texttt{APN}~(Ours) & &  &$63.1$  & $54.6$ & $30.5$ & $64.1$ & $55.9$ & $50.5$ & $52.8$ \\
\texttt{APN}~(Ours) & &  & $67.8$  & $55.9$ & $29.4$ & $68.7$ & $49.2$ & $49.1$ & $53.4$ \\

\hline
\texttt{SGMA}~\citep{SGMA} & \multirow{2}{*}{\xmark} & \multirow{2}{*}{$1/\sqrt{2}$} & $-$ & $-$  & $-$ & $74.9$ & $-$ & $48.1$ & $61.5$ \\
% \texttt{APN}~(Ours)  & &  & $88.9$ & $81.0$  & $72.1$ & $91.8$  & $76.6$ & $65.0$ & $79.2$\\
\texttt{APN}~(Ours)  & &  & $88.1$ & $81.3$  & $71.6$ & $91.4$  & $76.2$ & $70.8$ & $79.9$\\
%\bottomrule
\end{tabular}
}
\caption{\add{Body part localization in CUB dataset. We comparing our \texttt{APN} with detection model \texttt{SPDA-CNN} trained with part annotations~(row 2), and a ZSL model \texttt{SGMA}. The baseline \texttt{BaseMod} takes the original image feature $g(x)$ as input and is trained with $\mathcal{L}_{\Scale[0.6]{CLS}}$. For BB~(bounding box) size, $1/4$ means each part bounding box has the size $\frac{1}{4}W_b \times \frac{1}{4}H_b$ 
% \jimmy{use $\times$ instead of ``*''}
, where $W_b$ and $H_b$ are the width and height of the bird. 
We use gradient-based visual explanation method \texttt{CAM} to visualize the attribute attention map for the baseline \texttt{BaseMod}. 
% The first four methods are detection models trained with part annotation.
%  without part annotation. 
For a fair comparison, we use the same evaluation protocol as \texttt{SGMA} in the last two rows.} 
% \yongqin{can you highlight the best numbers in the table?}
}
\label{tab:atten_loc}
\end{table*}

% \begin{figure}[tb]
%   \centering
%     \includegraphics[width=1\linewidth]{IJCV/figures/CUB_quali_with_base.pdf}
%     \caption{Attribute localization (right): The color map covering the image indicates our attribute similarity maps. Red, blue, orange bounding boxes in the image represent the ground truth part bounding box, our results, and \texttt{BaseMod+CAM} respectively. Green~(purple) box outside the image indicates a correct~(incorrect) localization by our model. \zeynep{combine this figure and figure 2}}
%   \label{fig:att_add}
% \end{figure}

\subsection{Attribute and part localization in ZSL setting}

%As illustrated in Section~\ref{sec:Attri_loc}, our approach is capable of localizing different attributes in the image by inspecting the attention maps produced by the attribute prototypes. In this section, 
We first evaluate the part localization capability of our method quantitatively. We provide an ablation study and comparison with other methods in CUB~\citep{26_wah2011caltech}. In addition, we provide qualitative results of attribute localization on three benchmark datasets. Two user studies are performed to evaluate the accuracy and semantic consistency of attribute similarity maps.

\begin{figure*}[t]
  \centering
\includegraphics[width=\linewidth]{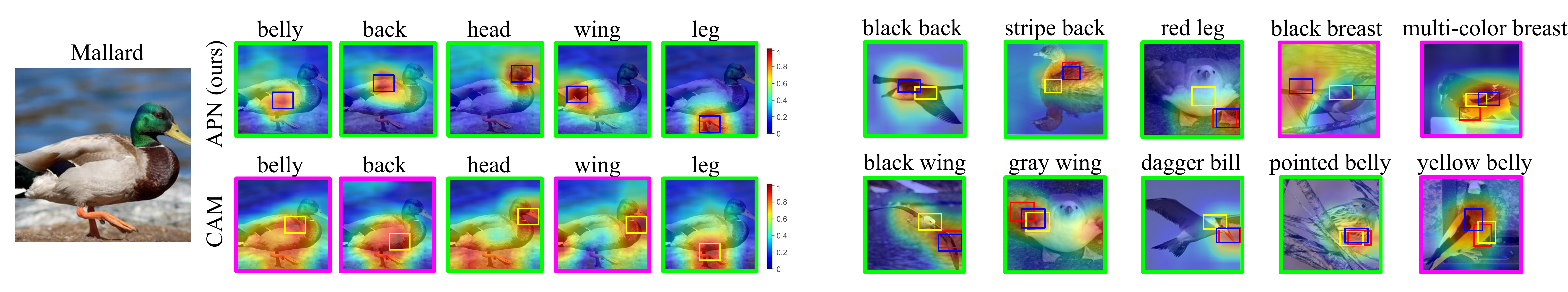}
    \caption{ Part and attribute localization on CUB.
        Left: 
        Attention maps for each body part of Mallard generated by our \texttt{APN}~(first row) and the baseline model \texttt{BaseMod} visualized by \texttt{CAM}~(\texttt{BaseMod(C)}, second row). Boxes mark out the area with the highest attention.
        Attention maps are min-max normalized for visualization. \add{Right: Various attribute similarity maps generated by our model. Red, blue, yellow bounding boxes in the image represent the ground truth part bounding box, our results, and the results of \texttt{BaseMod(C)} respectively.} Green~(purple) box outside the image indicates a correct~(incorrect) localization. 
        % \yongqin{Why not compare with Grad-CAM on CUB? It is compared on AWA and SUN}
        %\zeynep{adjust the spacing between images and sizes such that the left and the right figure occupies the same amount of space}%We cover the upsampled attention map on the original image, to show the corresponding location of the highlighted area. The bounding box marks out the most highlighted area. 
        % Compared to the \texttt{BaseMod}, we generate more accurate and concentrated maps.
    }
  \label{fig:all_part}
\end{figure*}

\subsubsection{Body part localization}

% \myparagraph{Evaluation protocol.} 
% As shown in Table~\ref{tab:attri_loc_def}, for each body part, there are several attribute subgroups encoded as $M$ sets of attribute indices $G_1,\dots, G_M$. For instance, \textit{breast} is related to two subgroups,  \textit{breast color} and \textit{breast pattern}, while \textit{breast color} subgroup consists of $15$ \jimmy{meaningless number could make reader confuse} color attributes such as \textit{black breast} and \textit{yellow breast}, etc.
% Over the $m$-th attribute subgroup, given the predicted value for each attribute $\{\hat{a}_k|k \in G_m\}$, we evaluate the similarity map for the attribute with the highest prediction score

% \begin{equation}
%     \argmax_{k \in G_m} \hat{a}_k\\.
% \end{equation}

% Thus, for each part, we average the part localization accuracy of $M$ similarity maps.
% \myparagraph{Evaluation protocol.} 
Given the attribute similarity maps related to six body parts of birds, we report the part localization accuracy by calculating the Percentage of Correctly Localized Parts~(PCP) following SGMA~\citep{SGMA}. 
As shown in Figure~\ref{fig:all_part}, the bounding box marks 
% \bernt{the word `out' does not make sense to me here - what you try to say? probably rephrase}
the image region with the highest attention in each attribute similarity map, then compared with the ground truth part annotation provided by the CUB~\citep{26_wah2011caltech} dataset.

\myparagraph{Ablation study.} 
We evaluates the effectiveness of our \texttt{APN} framework in terms of the influence of the attribute regression loss $\mathcal{L}_{\Scale[0.6]{Reg}}$, attribute decorrelation loss $\mathcal{L}_{\Scale[0.6]{AD}}$, the similarity compactness loss $\mathcal{L}_{\Scale[0.6]{CPT}}$, and the \texttt{ZoomInMod}. 
% Following SGMA~\citep{SGMA}, we report the part localization accuracy by calculating the Percentage of Correctly Localized Parts~(PCP). If the predicted bounding box for a part overlaps sufficiently with the ground truth bounding box, the detection is considered to be correct.
% The results are shown in Table~\ref{tab:ablation}~(right).
As shown in Table~\ref{tab:ablation}~(right), when trained with the joint losses, \texttt{APN} significantly improves the accuracy of \textit{breast}, \textit{head}, \textit{wing} and \textit{leg} by $27.5\%$, $44.5\%$, $13.2\%$, and $32.6\%$ respectively, while the accuracy of \textit{belly} and \textit{back} are improved less. This observation agrees with the qualitative results in Figure~\ref{fig:all_part} that \texttt{BaseMod} tends to focus on the center body of the bird, while \texttt{APN} results in more accurate and concentrated attention maps. 
Moreover, $\mathcal{L}_{\Scale[0.6]{AD}}$ boosts the localization accuracy, which highlights the importance of encouraging in-group similarity and between-group diversity when learning attribute prototypes.

\myparagraph{Comparing with SOTA.} 
We report PCP in Table~\ref{tab:atten_loc}.
As the baseline, we train a single \texttt{BaseMod} with cross-entropy loss $\mathcal{L}_{\Scale[0.6]{CLS}}$, 
% \jimmy{A smaller "CLS" like this.}
and use gradient-based visual explanation method \texttt{CAM}~\citep{CAM} to investigate the image area \texttt{BaseMod} used to predict each attribute. As state of the art, we report the part localization accuracy of a fine-grained classification model \texttt{SPDA-CNN}~\citep{spdacnn} which contains a bird part detection branch supervised by parts annotations. In the last two rows, we compare with the weakly supervised model~(without part annotation) \texttt{SGMA}~\citep{SGMA}, which learns part attention for \textit{head} and \textit{leg} by clustering feature channels.%\zeynep{explain the other methods in the table}

On average, our \texttt{APN} improves PCP over \texttt{BaseMod} by $22.7\%$~($53.4\%$ vs $30.7\%$). The majority of the improvements come from better leg and head localization, e.g. from $24.2\%$ to $68.7\%$~(head), and from $16.5\%$ to $49.1\%$~(leg). Compared to the supervised method \texttt{SPDA-CNN}, our method achieves comparable accuracy on \textit{breast} and \textit{wing}. Although there is still a gap to the supervised method \texttt{SPDA-CNN}, especially on the \textit{back} and \textit{leg} which are hard to localize, the results are encouraging since we do not need require part annotation during training.
% In the last two rows, we compare with the weakly supervised \texttt{SGMA}~\citep{SGMA} model which learns part attention by clustering feature channels.
Since the feature channels encode more pattern information rather than local information~\citep{geirhos2018imagenet,zhou2018interpreting}, enforcing locality over spatial dimension is more accurate than over channel. Under the same evaluation protocol, we significantly improve the localization accuracy over SGMA~($79.9\%$ vs $61.5\%$ on average). Besides, our model is able to localize attributes related to six body parts, while SGMA can only localize two main parts.

\subsubsection{Qualitative results}
%\zeynep{Write 2 sentences explaining the experiments.}
Our model localizes attributes by inspecting the attention maps produced by the attribute prototypes. In this section, we qualitatively investigate the part localization ability on the CUB dataset, as well as the attribute localization results on three benchmark datasets.

\myparagraph{Part localization in CUB.}
We first investigate the difference between our \texttt{APN} and the baseline \texttt{BaseMod} for localizing different body parts in CUB dataset. In Figure~\ref{fig:all_part} (left), for each part of the bird Mallard, we display one attribute similarity map generated by our model \texttt{APN}, and the baseline model \texttt{BaseMod} visualized by \texttt{CAM}~(\texttt{BaseMod(C)}). The baseline model tends to generate disperse attention maps covering the whole bird, as it utilizes more global information, e.g. correlated bird parts and context, to predict attributes. For instance, when predicting attributes of \textit{belly} and \textit{back}, the baseline model utilizes pixels scattered on the bird body.
% as it utilizes the global information extracted from the image to predict attributes. 
On the other hand, the similarity maps of our \texttt{APN} are more concentrated and diverse and therefore they localize different bird body parts more accurately. The improvement is lead by the attribute prototypes and compactness loss, which helps the model to focus on local image features when learning attributes. 
% Following SGMA~\citep{SGMA}, we use a bounding box to ground the most highlighted area as the detected attribute patch.

\myparagraph{Attribute localization in CUB.}
\add{In addition, unlike other models~\citep{spdacnn,uijlings2013selective,SGMA} that can only localize body parts, our \texttt{APN} model can provide attribute-level localization for CUB dataset, as shown in Figure~\ref{fig:all_part} (right). Our model can localize back with various shapes ~(row 1, column 1 and 2) and wings with different postures ~(row 2, column 1 and 2). We can also locate the \textit{pointed belly} of the occluded bird~(row 2, column 4). Compared with \texttt{BaseMod(C)}, our approach produces more accurate bounding boxes that localize the predicted attributes.
% , demonstrating the effectiveness of our attribute prototype network. 
For example, while \texttt{BaseMod(C)} wrongly learns the \textit{red leg} from the image region of the belly~(row 1, column 3), our model precisely localizes the \textit{red leg} at the correct region. Specifically, when predicting the attributes for wings and legs, the model tends to focus on both two wings and legs even when they are physically separated~(row 1, column 3 for \textit{red leg} and row 2, column 2 for \textit{gray wing}). These results are interesting because our model is trained on only class-level attributes without accessing any bounding box annotation. }

\add{As a side benefit, the attribute localization ability introduces a certain level of interpretability that supports the zero-shot inference with attribute-level visual evidence. 
The last two columns in Figure~\ref{fig:all_part} (right) show some failure examples where our model makes wrong predictions. For example, when the yellow breast is wrongly predicted as black~(row 1, column 4), the attention map tends to spread over the tail and background; when the black breast is recognized as multi-colored~(row 1, column 5), the attention map points to the region of the black and white wing.
% We observe that our predicted bounding boxes for the \textit{gray wing} and \textit{blue forehead}~(row 5) are not completely wrong while they are considered as failure cases by the evaluation protocol. 
Besides, although our attribute decorrelation loss in Equation~\ref{eq:ad} alleviates the correlation issue to some extent~(as shown in the previous results in Table 1 and Figure 2), we observe that our \texttt{APN} seems to still conflate the \textit{yellow belly} and \textit{yellow breast}~(row 2, column 5) in some cases, indicating the attribute correlation issue as a challenging problem for future research. }

\begin{figure*}[t]
  \centering
  \includegraphics[width=0.9\linewidth]{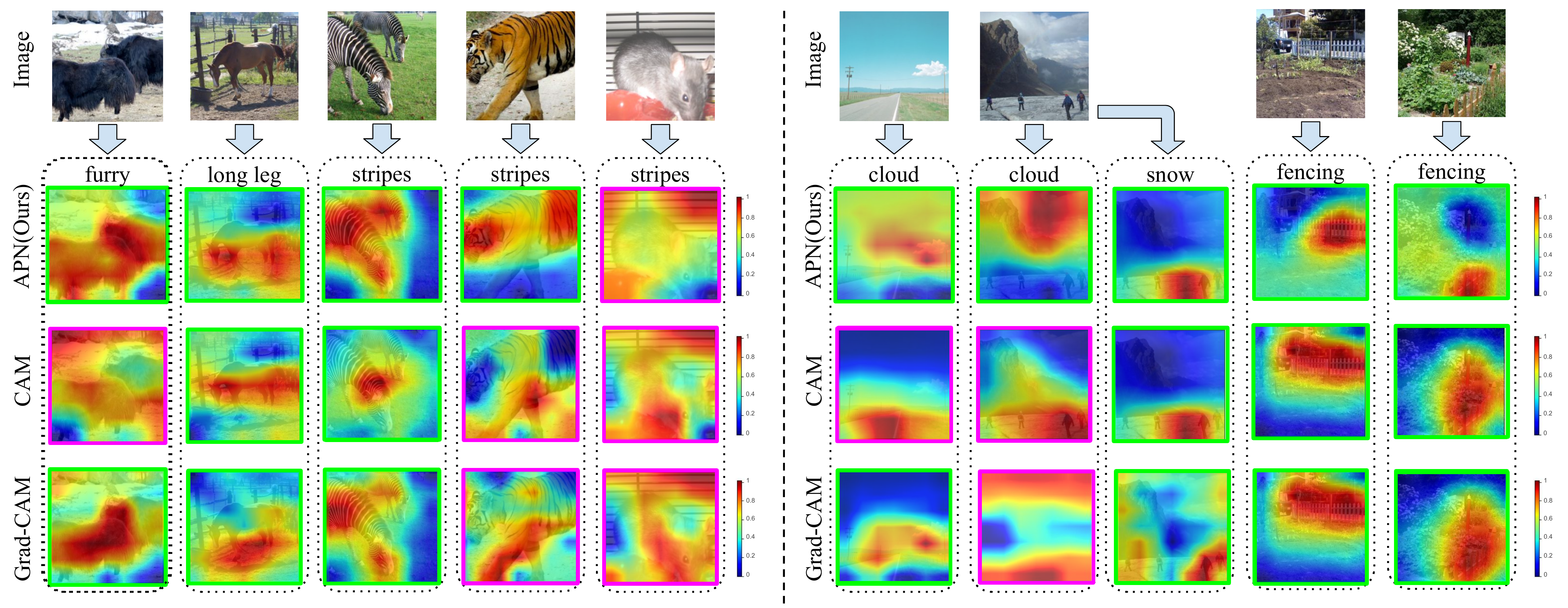}
    \caption{\add{Attribute attention maps for AWA2~(left) and SUN~(right) from our \texttt{APN} model~(row 2), \texttt{BaseMod} visualized with \texttt{CAM}~(row 3), and \texttt{BaseMod} with \texttt{Grad-CAM}~(row 4). The attribute similarity maps are min-max normalized for visualization. We cover the upsampled attention map on the original image, to show the corresponding location of the highlighted area. %The caption above the attribute attention maps indicates the attribute name. 
    Green~(purple) box outside the image indicates a correct~(incorrect) localization.}
    % The purple box outside the image indicates an incorrect localization. 
    % \jimmy{use ``SimMap'' or ``Sim-map'' instead of ``Sim\_map''}
    }
\label{fig:SUN_AWA_quali}
\end{figure*}

\myparagraph{Attribute localization in AWA2 and SUN.}
% We also investigate if our attribute prototype network can localize visual attributes on AWA2 dataset.
\add{To show how these observations generalize in two other datasets, in Figure~\ref{fig:SUN_AWA_quali}, we compare our \texttt{APN} model with two baseline models on AWA2 and SUN. The attribute attention maps of \texttt{BaseMod} is generated with two gradient-based visual explanation method \texttt{CAM}~\citep{CAM} and \texttt{Grad-CAM}~\citep{grad_cam}.} 

\add{In AWA2 dataset~(Figure~\ref{fig:SUN_AWA_quali}, left), our network produces precise similarity maps for visual attributes that describe \textit{texture} and \textit{body parts}, etc.
We can localize visual attributes with diverse appearances, e.g. the white and black \textit{stripe} of zebra, and the yellow and black \textit{stripe} of tiger~(row 2, column 3,4), while 
\texttt{CAM} and \texttt{Grad-CAM} fails in localizing the stripe on tigers.
Our similarity maps for \textit{furry} and \textit{longleg} can precisely mask out the image regions of the ox and horse (row 2, column 3,4), while \texttt{BaseMod} only localizes part of the image~(row 1, column 3,4).
% On the other hand, there are some failure cases which are marked in purple bounding box. 
On AWA2 dataset we are interested in the visual attributes of animals, while our model in some cases highlights the attributes of the background, e.g. identifying the grid on the rat cage as stripes~(row 2, column 5). This can be explained by the fact that our model only relies on weak supervision, i.e. class-level attributes and their semantic relatedness.}

\add{The attribute similarity maps on SUN dataset are shown in figure~\ref{fig:SUN_AWA_quali}~(right). Our model can discriminate between different attributes with similar color or texture, e.g. correctly locating \textit{snow} and \textit{cloud} in one image~(row 2, column 2,3). The baseline models, on the other hand, cannot distinguish between cloud and snow. Although the appearance of one visual attribute may vary significantly, we can still locate them correctly, e.g. the \textit{fencing} with different colors, location, and shape~(row 2, column 4,5). 
Overall, those results indicate that we can perform attribute localization in a weakly supervised manner and provide visual evidence for the inference process of ZSL.}

\begin{figure*}[t]
  \centering
  \includegraphics[width=0.4\linewidth]{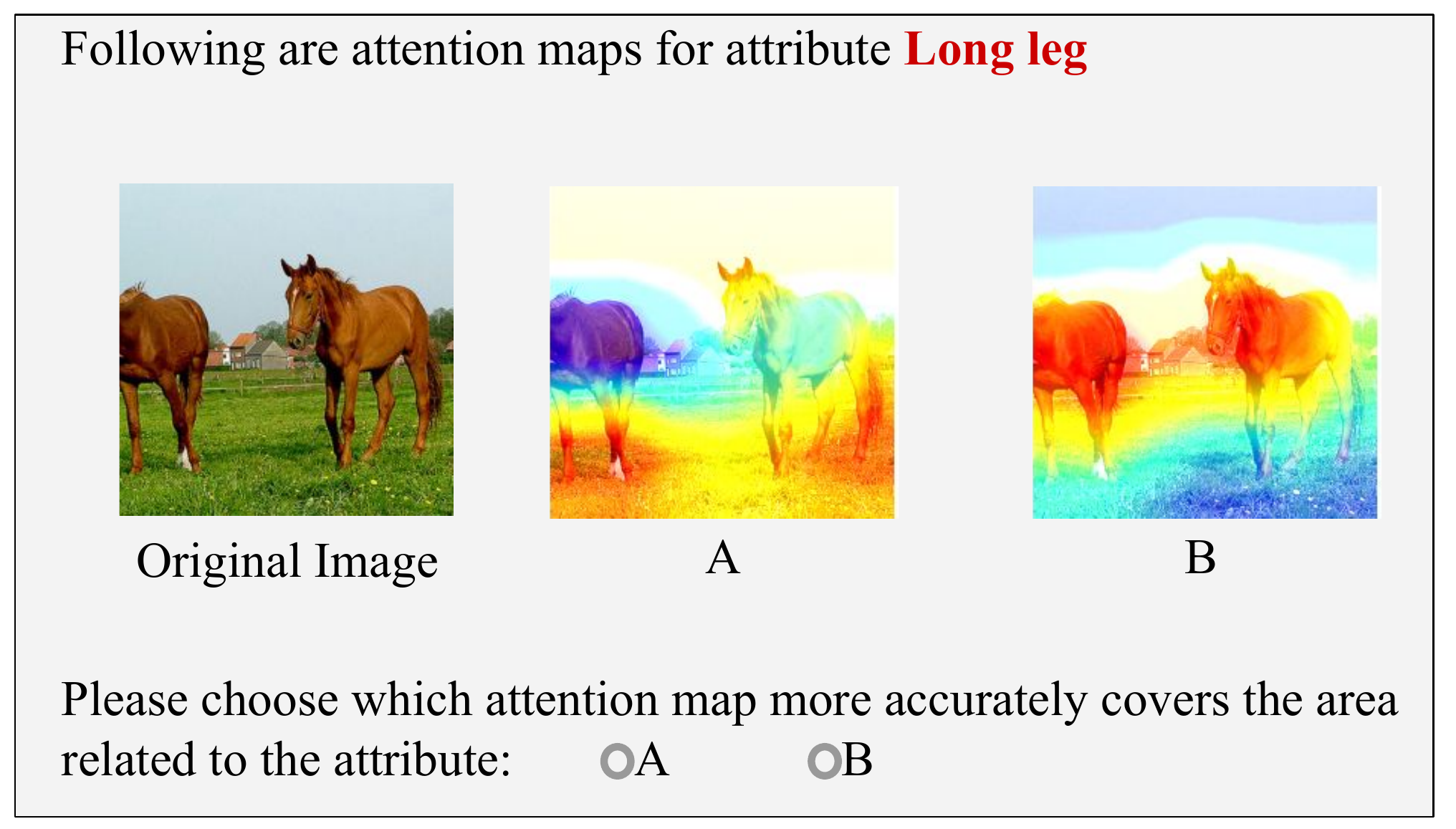} %\hfill
  \includegraphics[width=0.4\linewidth]{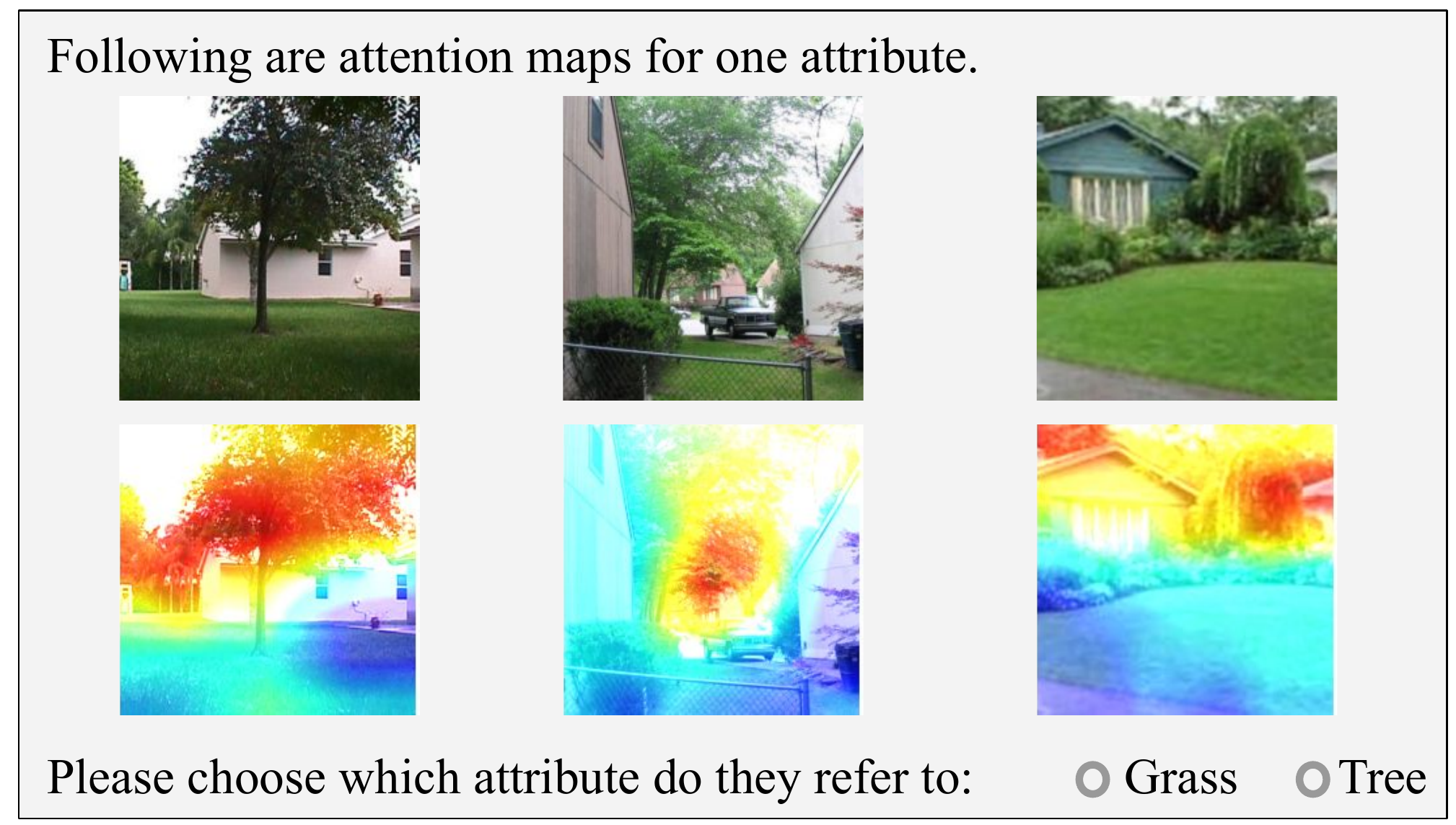}
    \caption{\add{User Study. Left: User study interface to evaluate the accuracy. Participants are required to choose the attention map that more accurately covers the attribute related region from two attention maps generated by our \texttt{APN} model and the \texttt{BaseMod}~(in random position). %\zeynep{give more details. the caption is too vague}. 
    Right: User study interface for the semantic consistency. Given three attention maps generated for an attribute, the user selects from two candidate attributes which one is the target of the attention maps.} %\zeynep{give more details. the caption is too vague}
    % Human annotators are asked to choose the attention map that covers the attribute-related region more accurately.
    }
\label{fig:Userstudy}
\end{figure*}

{
\setlength{\tabcolsep}{4pt}
\renewcommand{\arraystretch}{1.2} 
\begin{table}[t]
\centering
\resizebox{\linewidth}{!}{%
    \centering
    \begin{tabular}{l l | c }
    %\hline
    %& \multicolumn{2}{c|}{Accuracy} & \multicolumn{3}{c}{Semantic consistency}\\ %\hline
    & \textbf{Methods} & \textbf{Results} \\ \hline
    Accuracy &  \texttt{APN} vs \texttt{CAM} & $\textbf{76.0}\%$ vs $24.0\%$~($\pm 5.1\%$) \\
     & \texttt{APN} vs \texttt{Grad-CAM} & $\textbf{74.4}\%$ vs $25.6\%$~($\pm 2.9\%$) \\ \hline 
    Semantic  & \texttt{CAM} & $55.0\%$~($\pm 3.2\%$)\\
    consistency  & \texttt{Grad-CAM} & $60.0\%$~($\pm 5.5\%$) \\
      & \texttt{APN} & $\textbf{89.0}\%$~($\pm 3.7\%$) \\
      %\hline
    \end{tabular}
    }
    \caption{\add{User study results. Top: the percentage of time that one method is marked as more accurate than the other one by users. Bottom: the percentage of attention maps that can be correctly associated with the target attribute by users. \texttt{APN}, \texttt{CAM}, and \texttt{Grad-CAM} denote our model, baseline model visualized by \texttt{CAM}~\citep{CAM}, and \texttt{Grad-CAM}~\citep{grad_cam} respectively.}}
    \label{tab:Userstudy}
\end{table}
}

\subsubsection{User study}

% \begin{table}[tb]
% \centering
% \resizebox{\linewidth}{!}{%
%     \centering
%     \begin{tabular}{c|c|c|c|c}
%     \hline
%     \multicolumn{2}{c|}{Accuracy} & \multicolumn{3}{c}{Semantic consistency}\\\hline
%       APN vs CAM  & APN vs Grad-CAM & CAM & Grad-CAM & APN \\
%       \hline
%       $76.0\%$ vs $24.0\%$~($\pm 5.1\%$)  & $74.4\%$ vs $26.0\%$~($\pm 2.9\%$) & $55.0\%$~($\pm 3.2\%$) & $60.0\%$~($\pm 5.5\%$)&   $89.0\%$~($\pm 3.7\%$)\\
%       \hline
%     \end{tabular}
%     }
%     \caption{Userstudy results. Left: the percentage of time being marked as more accurate than the other by users. Right: the percentage of attribute attention maps being learnable~(can be correctly recognized) by humans.\wenjia{I will split the table into two subtables(top and bottom).}}
%     \label{tab:Userstudy}
% \end{table}

\add{Since the CUB dataset is the only one among the datasets considered here that contains ground truth parts, we design two user studies to assess the accuracy and semantic consistency of attribute attention maps from SUN and AWA2 datasets. We compare the performance of our \texttt{APN} model and the baseline \texttt{BaseMod} visualized by two model explanation methods, \texttt{Grad-CAM} and \texttt{CAM}, respectively.}

\myparagraph{Accuracy of attribute localization.} 
\add{The goal is to evaluate whether the attribute attention maps precisely attend to the related image area. As shown in Figure~\ref{fig:Userstudy} (left), each test is a tuple $\left ( M^{i}_{\Scale[0.6]{APN}}, M^{i}_{\Scale[0.6]{BaseMod}}, a_i\right )$ for attribute $a_i$, where $M^{i}_{\Scale[0.6]{APN}}$ is the attribute attention maps generated by our \texttt{APN} model, and  $M^{i}_{\Scale[0.6]{BaseMod}}$ by the \texttt{BaseMod}. The human annotators are presented with the tuple
% the attribute and two attention maps,
and they are asked to choose the attention map that more accurately covers the attribute region. We randomly sample 50 attention maps from our \texttt{APN} model for 20 visual attributes, then generate the corresponding \texttt{Grad-CAM} and \texttt{CAM} attention maps for \texttt{BaseMod}, and create $100$ tuples in total. Two separate experiments are performed to compare our \texttt{APN} model with the \texttt{BaseMod} visualized by \texttt{Grad-CAM} and \texttt{CAM}.
For each experiment, we employed 5 annotators, i.e. in total 10 students (4 female) aged between 20 and 30 and majoring in computer science participated in the experiment. }

\add{We average the responses from each participant and report the standard deviation between each participant as well as the overall accuracy in Table~\ref{tab:Userstudy}~(top). As for the result, our \texttt{APN} attribute attention maps outperform the \texttt{BaseMod} by a large margin.  When comparing the attribute attention maps generated by our \texttt{APN} model and the \texttt{BaseMod} visualized by \texttt{CAM}, in $76.0\%$ cases, \texttt{APN} are marked as more accurately covers the attribute-related area than \texttt{CAM}. And in $74.4\%$ cases, $\texttt{APN}$ are more accurate than \texttt{Grad-CAM}. The user study results agree with the qualitative results in Figure~\ref{fig:SUN_AWA_quali} that \texttt{APN} demonstrates more accurate attribute attention maps than the baseline model. }
%\zeynep{do you average the responses of the participants. Much of the details of this experiment is missing. Please extend this part more.}
 %While in only $24.0\%$ cases, users vote for the baseline model. 
% The statistical significance are verified by Chi-Square test with $p<10^{-5}$. \zeynep{you can not run Chi-square test for 5 participants. What do you mean here? You can provide standard deviation in a table instead.}
%\zeynep{provide a table for these results}

% \begin{figure*}[t]
%   \centering
%     \includegraphics[width=\linewidth]{IJCV/figures/FSL_all_3.pdf}
%     \caption{Few-shot learning~(top) and generalized few-shot learning~(bottom) results on CUB, AWA2 and SUN datasets. We apply our \textit{APN features} to feature generation model~(i.e.  \texttt{APN+f-VAEGAN-D2}) and compare with other data synthesis based models. For fair comparison, models marked with $^*$~(e.g. \texttt{Analogy$^*$}) use \textit{finetuned features} extracted from ResNet101. FSL plots show the top-1 accuracy on novel classes, and GZSL plots show the top-1 accuracy on all classes.}
%   \label{fig:FSL_all}
% \end{figure*}

\myparagraph{Semantic consistency.} 
\add{Here, our aim is to measure whether the attention maps on different images for one attribute is semantically consistent and can be understood by human. 
% Formally, we conduct a number of human-based tests to assess the semantic consistency. 
Each test is a tuple $\left ( \mathcal{M}^{i}, a_i, a_j \right )$, where $a_i$ is the target attribute, and $a_j$ is a distractor attribute that is semantically similar to $a_i$. $\mathcal{M}^{i} = \left \{ M^{i}_1, M^{i}_2, M^{i}_3\right \}$ is a random subset of attribute attention maps for the target attribute $a_i$. As shown in Figure~\ref{fig:Userstudy} (right), the human annotators are presented with the tuple via a user interface,
% three sampled attention maps, as well as the target and the distractor attribute via a user interface, 
and their task is to identify which of the attributes does the attention map refer to. The performance is defined as the average accuracy of solving such tasks correctly.
We sample $20$ tuples for three methods~(\texttt{APN}, \texttt{BaseMod+CAM} and \texttt{BaseMod+Grad-CAM}) with two criterion: 1) the evaluated original image and attributes stay the same for the three methods, 2) the target $a_0$ and distractor $a_1$ attributes all appear in the original images. Three experiments are performed to evaluate the three methods. There are five annotators for each test, and in total 10 students (4 female) aged between 20 and 30 and majoring in computer science participated in the experiment.}
%\zeynep{provide the demographics as before}
% For each test, there are five annotators, and over 300 human-based tests are performed. 

% We conduct experiments to verify the semantical consistency for attention maps from \texttt{APN}  and \texttt{BaseMod} separately. 
\add{We average the responses from each user and report their overall accuracy and standard deviation between each user in Table~\ref{tab:Userstudy}~(bottom).
The user study results indicate high semantic consistency for our model. \texttt{APN} attention maps can be associated with the correct attribute in $89.0\%$ cases, while attention maps from \texttt{BaseMod+CAM} only achieve $55.0\%$ accuracy. \texttt{BaseMod+Grad-CAM} gains a higher result, at $60.0\%$, but is still $29.0\%$ lower than ours. 
% The statistical significance is verified by the 2-sample z-test on pro-portions with $p<0.01$. 
The results indicate that our model is able to generate attribute attention maps that are semantic coherent, and can be understood by human. Although the distractor attribute $a_0$ shares a very similar semantic meaning with the correct attribute~(as shown in Figure~\ref{fig:Userstudy}~(right)), \texttt{APN} attention maps can still help users to find the correct answer. }
%\zeynep{same comments as above in the accuracy of attribute localization paragraph. please implement the same suggestions here.}

%----------FSL--------------

\subsection{Few-shot learning}

\new{In the few-shot learning~(FSL) scenario, the images are divided into base classes where plenty of training samples can be obtained and novel classes with only a handful of training samples. The goal of FSL is to learn a classifier to recognize novel classes with limited labeled examples. In the generalized few-shot learning~(GFSL) setting, the classifier is trained to recognize images from both base and novel classes.}

% In this section, we choose to verify our model in a more realistic task, the many-way few-shot learning problem, that aims to identify many categories at the same time.}
% where each category has only a few samples.

\begin{table*}[t]
\centering
\resizebox{\linewidth}{!}{%
    \begin{tabular}{c | c c c c | c c c c | c c c c}
    % \toprule
        \multirow{2}{*}{\textbf{Method}} & \multicolumn{4}{c|}{\textbf{AWA2}}  & \multicolumn{4}{c}{\textbf{CUB}} & \multicolumn{4}{c}{\textbf{SUN}} 
        \\ 
        & \textbf{1} & \textbf{2} & \textbf{5} & \textbf{10} & \textbf{1} & \textbf{2} & \textbf{5} & \textbf{10} & \textbf{1} & \textbf{2} & \textbf{5} & \textbf{10} \\ \hline
\texttt{Analogy}~\citep{hallucinate2017low}      &       62.5 & 81.3 & 82.4 & 87.8 & 56.5  & 69.5  & 78.0   & 81.1   & 40.0    & 49.0    & 67.6 & 75.5 \\
\texttt{Imprinted}~\citep{qi2018low}     &     66.9 & 82.6 & 89.0   & 93.5 & 48.5  & 65.0    & 80.0   & 85.3 & 37.9  & 47.8  & 62.1 & 70.2 \\  \hline
\texttt{f-VAEGAN-D2}~\citep{xian2019}    &   75.0   & 87.9 & 90.5 & 93.1 & 76.1  & 79.6  & 83.4 & 85.9 & \textbf{68.8}  & 69.5  & 70.0   & 70.9 \\
\texttt{APN+f-VAEGAN-D2}~(Ours)       &     82.1 & 92.4 & 94.2 & 95.6 & \textbf{77.8}  & 81.1  & 84.8 & \textbf{87.1} & 68.2  & 69.4  & 72.4 & 75.1 \\ \hline
\texttt{TF-VAEGAN}~\citep{tfvaegan} & 77.3 & 84.4 & 87.7 & 89.8 & 75.6  & 81.1  & 83.5 & 85.6 & 68.1  & 68.5  & 68.9 & 74.0   \\
\texttt{APN+TF-VAEGAN}~(Ours)  &   \textbf{87.2} & \textbf{92.8} & \textbf{94.7} & \textbf{96.0}   & 77.1  & \textbf{82.6}  & \textbf{85.2} & 86.6 & \textbf{68.8}  & \textbf{69.9}  & \textbf{74.1} & \textbf{76.0}  \\
      
    % \bottomrule
    \end{tabular}}
    \caption{\new{Few-shot learning results. We apply our \textit{APN features} to feature generation model~(i.e.  \texttt{APN+f-VAEGAN-D2} and \texttt{APN+TF-VAEGAN}) and compare with other data synthesis based models. }
    \revise{Note that \texttt{f-VAEGAN-D2}~\citep{xian2019} and \texttt{TF-VAEGAN}~\citep{tfvaegan} generate image features given the class attributes to augment the training of novel classes. Our \texttt{APN} model uses the same attributes to learn locality enhanced image features.}}
    \label{tab:fsl}
\end{table*}

\begin{table*}[t]
\centering
\resizebox{\linewidth}{!}{%
    \begin{tabular}{c | c c c c | c c c c | c c c c}
    % \toprule
        \multirow{2}{*}{\textbf{Method}} & \multicolumn{4}{c|}{\textbf{AWA2}}  & \multicolumn{4}{c}{\textbf{CUB}} & \multicolumn{4}{c}{\textbf{SUN}} 
        \\ 
        & \textbf{1} & \textbf{2} & \textbf{5} & \textbf{10} & \textbf{1} & \textbf{2} & \textbf{5} & \textbf{10} & \textbf{1} & \textbf{2} & \textbf{5} & \textbf{10} \\ \hline
\texttt{Analogy}~\citep{hallucinate2017low}      &    55.0   & 64.7 & 70.7 & 74.5 & 54.1    & 67.5  & 75.5  & 79.5 & 37.5  & 39.6  & 42.4  & 44.6 \\
\texttt{Imprinted}~\citep{qi2018low}     &  44.7 & 50.5 & 70.0   & 88.1 & 57.0    & 67.5  & 75.5  & 79.5 & 37.7  & 38.9  & 42.4  & 42.5 \\  \hline
\texttt{f-VAEGAN-D2}~\citep{xian2019}    & 72.7 & 80.7 & 85.6 & 88.9 & 71.1  & 74.5  & 77.6  & 79.7 & 43.5  & 43.5  & 43.2  & 44.9 \\
\texttt{APN+f-VAEGAN-D2}~(Ours)   &   74.6 & 83.3 & 86.9 & 89.7 & \textbf{72.1}  & 76.1  & \textbf{79.3}  & \textbf{80.9} & 43.1  & \textbf{44.0}    & 45.5  & 45.9 \\ \hline
\texttt{TF-VAEGAN}~\citep{tfvaegan} & 74.7 & 81.1 & 86.0   & 87.1   & 70.6  & 74.7  & 77.5  & 79.3 & 43.3  & 42.3  & 42.9  & 45.4 \\
\texttt{APN+TF-VAEGAN}~(Ours)  & \textbf{80.1} & \textbf{85.9} & \textbf{88.6} & \textbf{90.3} & 71.7  & \textbf{76.4}  & 78.9  & 80.4 & \textbf{44.3}  & 42.9  & \textbf{46.1}  & \textbf{46.7} \\
      
    % \bottomrule
    \end{tabular}}
    \caption{\new{Generalized few-shot learning results. We apply our \textit{APN features} to feature generation model~(i.e.  \texttt{APN+f-VAEGAN-D2} and \texttt{APN+TF-VAEGAN}) and compare with other data synthesis based models. }}
    \label{tab:gfsl}
\end{table*}

\myparagraph{Compared methods.} 
\new{In this section, we evaluate our attribute prototype network under two evaluation protocols, i.e. the all-way evaluation and N-way-K-shot evaluation.
In the all-way evaluation, 
% to perform few-shot learning, we train the attribute prototype network on base classes, then apply our \textit{APN feature} to generative models \texttt{f-VAEGAN-D2}~\citep{xian2019} and \texttt{TF-VAEGAN}~\citep{tfvaegan}, and train a FSL classifier with the generated samples. O
our model is compared with several state-of-the-art generative FSL methods.  \texttt{Analogy}~\citep{hallucinate2017low}, \texttt{f-VAEGAN-D2}~\citep{xian2019} and \texttt{TF-VAEGAN}~\citep{tfvaegan} are data synthesis based methods that augment image features for novel classes. \texttt{Imprinted}~\citep{qi2018low} directly uses the normalized activation of novel images as the classifier weight. For a fair comparison, these models are trained with \textit{finetuned feature} extracted from ResNet101, and under the same dataset split~\citep{xian2019}. In the FSL setting, we report the averaged top-1 accuracy for novel classes with the model that only predicts novel classes. In the GFSL setting, we report the averaged top-1 accuracy of test samples of all classes, where the model predicts both base and novel classes labels.}

\new{In the N-way-K-shot evaluation, we build our attribute prototype network over the ResNet12~\citep{he2016deep} backbone in \texttt{DPGN}~\citep{yang2020dpgn}, and train the network with both the FSL training losses~\citep{yang2020dpgn} and our $\mathcal{L}_{\Scale[0.6]{APN}}$ loss. Our model is compared with state-of-the-art FSL methods under the N-way-K-shot evaluation. \texttt{MatchingNet}~\citep{vinyals2016matching}, \texttt{ProtoNet}~\citep{snell2017prototypical} and \texttt{CloserLook}~\citep{fu2017look} propose to optimize the representation learning model with metric learning methods. \texttt{MAML}~\citep{finn2017} learns to initialize the model weight so that it can adapt to novel classes efficiently. Two latest methods further enhance the meta-learning approach with graph network~\citep{yang2020dpgn} and attention agent trained with reinforcement learning~\citep{hong2021reinforced}. We report the top-1 accuracy in the 5way-1shot and 5way-5shot setting following~\citep{yang2020dpgn}.}

\myparagraph{Comparing with the SOTA.} 
\new{We display the few-shot learning accuracy in Table~\ref{tab:fsl}. Under the all-way evaluation setting, our model yields consistent improvement on three datasets, i.e. CUB, AWA2, and SUN. On AWA2 dataset, compared to \texttt{TF-VAEGAN} trained with \textit{finetuned feature}, our model \texttt{APN+TF-VAEGAN} improve the FSL accuracy of by a large margin, especially in the low-shot scenario where only a small number of samples from novel classes are available, i.e. we gain $9.9\%$~(1-shot), $8.4\%$~(2-shot), $7.0\%$~(5-shot), and $6.2\%$~(10-shot). }
\add{On fine-grained dataset CUB, our \textit{APN feature} yields consistent improvement over the \textit{finetuned feature}. Compared to \texttt{f-VAEGAN-D2} trained with \textit{finetuned feature}, our model \texttt{APN+f-VAEGAN-D2} gain $1.7\%$~(1-shot), $1.5\%$~(2-shot), $1.4\%$~(5-shot), and $1.2\%$~(10-shot) on FSL. The same trend is observed on SUN dataset. Though the accuracy for the 1-shot and 2-shot regime are comparable to other methods, we manage to improve a lot when training samples increase, e.g. our model \texttt{APN+TF-VAEGAN} achieves $74.1\%$~(5-shot) and $76.0\%$~(10-shot), compared to \texttt{TF-VAEGAN} with $68.9\%$~(5-shot) and $74.0\%$~(10-shot).}

\new{Compared to other FSL models, we gain significant improvements. For instance, we achieve $77.8\%$~(1-shot) on FSL for CUB dataset, compared to $48.5\%$~(\texttt{Analogy}) and $56.5\%$~(\texttt{Imprinted}). For AWA2 dataset, we gain $87.2\%$ on FSL, compared to $66.9\%$~(\texttt{Analogy}) and $62.5\%$~(\texttt{Imprinted}). Specifically, our model using one labeled image per class approaches the accuracy of \texttt{Analogy} and \texttt{Imprinted} trained with five samples. It indicates that improving the locality of image features can better help the feature generators to mimic the real data distribution. When increasing the number of training samples~(i.e. in the 10-shot scenario), the distance between the other three methods shrinks as we are going towards the fully supervised setting. However, our model still manages to improve the performance, which denotes that even with abundant training samples, locality enhanced image features will train a more discriminative classifier than ordinary features.}

\new{The locality augmented model generates discriminative features for novel classes, especially when applied to the generalized setting where the model should predict both base and novel classes. As shown in Table~\ref{tab:gfsl}, on AWA2 dataset, our model $\texttt{APN+TF-VAEGAN}$ gains accuracy improvement for $5.4\%$~(1-shot), $4.8\%$~(2-shot), $2.6\%$~(5-shot), and $3.2\%$~(10-shot). The results demonstrate that our model generates highly discriminative image features by leveraging attribute information. In CUB dataset, we gain $1.0\%$~(1-shot), $1.6\%$~(2-shot), $1.7\%$~(5-shot), and $1.1\%$~(10-shot). In SUN dataset, we also manage to improve the accuracy consistently, e.g. we gain $1.0\%$~(1-shot), $0.7\%$~(2-shot) and $3.2\%$~(5-shot), and $1.3\%$~(10-shot) on GFSL.}

\begin{table}[t]
    \centering
    \resizebox{\linewidth}{!}{%
    \begin{tabular}{c |c |c c}
        \textbf{Method} & \textbf{Backbone} & \textbf{5way-1shot} & \textbf{5way-5shot}  \\
    \hline
        % \texttt{MatchingNet} & ConvNet & $43.56\pm0.84$ & $55.31\pm0.73$ \\
        % \texttt{ProtoNet} & ConvNet & $51.31\pm0.91$ & $70.77\pm0.69$ \\
        \texttt{MatchingNet*} & ResNet18 & $72.4\pm0.90$ & $83.6\pm0.60$ \\

        \texttt{ProtoNet*} & ResNet18 & $73.0\pm0.88$ & $86.6\pm0.51$ \\

        \texttt{MAML*} & ResNet18 & $68.4\pm1.07$ & $83.5\pm0.62$ \\
        \texttt{CloserLook} & ConvNet & $60.5\pm0.83$ & $79.3\pm0.61$ \\
        \texttt{ArL} & ConvNet & $50.6$ & $65.9$ \\
        % \texttt{FEAT}~\citep{chen2019closer} & ResNet12 & $68.87\pm0.22$ & $82.90\pm0.15$ \\
        \texttt{RAP+ProtoNet} & ResNet18 & $74.1\pm0.60$ & $89.2\pm0.31$ \\ 
        \texttt{RAP+Neg-Margin} & ResNet18 & $75.4\pm0.81$ & $90.6\pm0.39$ \\ \hline
        \texttt{DPGN} & ResNet12 & $75.7\pm0.47$ & $91.5\pm0.33$ \\
        \texttt{APN+DPGN} (Ours) & ResNet12 & $\textbf{77.4}\pm0.44$ & $\textbf{92.2}\pm0.24$  \\ 
        % RAP+LaplacianShot(CVPR2021) & ResNet18 & $83.59\pm0.18$ & $90.77\pm0.10$ \\
        % \texttt{LaplacianShot} (ICML-20)& ResNet18 & $80.96$ & $88.68$ \\
        % \texttt{APN+LaplacianShot} (Ours) & ResNet18 & - & - \\

    \end{tabular}}
    \caption{\new{Few-shot learning results on CUB dataset (* results are from~\citet{hong2021reinforced}). Following~\citet{yang2020dpgn}, we report top-1 accuracy in 5way-1shot/-5shot settings and compare ours with \texttt{MatchingNet}~\citep{vinyals2016matching}, \texttt{ProtoNet}~\citep{snell2017prototypical}, \texttt{MAML}~\citep{finn2017}, \texttt{CloserLook}~\citep{fu2017look},  \texttt{ArL}~\citep{zhang2021rethinking}, \texttt{RAP+ProtoNet}~\citep{hong2021reinforced}, \texttt{RAP+Neg-Margin}~\citep{hong2021reinforced}, \texttt{DPGN}~\citep{yang2020dpgn}. }}
    \label{tab:meta_cub}
\end{table}

\setlength{\tabcolsep}{4pt}
\renewcommand{\arraystretch}{1.2} 
\begin{table*}[h]
\centering
\small
\resizebox{\linewidth}{!}
{\begin{tabular}{l| x{1.2cm} x{1.2cm} x{1.2cm} |c c c |c c c |c c c}
   %\toprule
      & \multicolumn{3}{c|}{\textbf{Zero-Shot Learning (ZSL)}} & \multicolumn{9}{c}{\textbf{Generalized Zero-Shot Learning (GZSL)}} \\
      & \textbf{AWA2} & \textbf{CUB} & \textbf{SUN} & \multicolumn{3}{c}{\textbf{AWA2}} & \multicolumn{3}{c}{\textbf{CUB}} & \multicolumn{3}{c}{\textbf{SUN}}  \\
     Method & \textbf{T1} & \textbf{T1} & \textbf{T1} &  \textbf{u} & \textbf{s} & \textbf{H} & \textbf{u} & \textbf{s} & \textbf{H} & \textbf{u} & \textbf{s} & \textbf{H}\\
    \hline

    \texttt{CADA-VAE}~\citep{schonfeld2019} & $49.0$  & $22.5$  & $\textbf{37.8}$  & $38.6$  & $60.1$  & $47.0$  & $16.3$  & $39.7$  & $23.1$  & $26.0$  & $28.2$ & $\textbf{27.0}$ \\
    \texttt{SJE}~\citep{akata2015evaluation} & $53.7$  & $14.4$  & $26.3$  & $39.7$  & $65.3$  & $48.8$  & $13.2$  & $28.6$  & $18.0$  & $19.8$  & $18.6$ & $19.2$ \\
    \texttt{GEM-ZSL}~\citep{liu2021goal} & $50.2$  & $25.7$  & $-$  & $40.1$  & $80.0$  & $53.4$  & $11.2$  & $48.8$  & $ 18.2$  & $-$  & $-$ & $-$  \\
    \texttt{APN(w2v)}~(Ours) & $\textbf{59.6}$  & $\textbf{27.7}$  & $32.1$  & $41.8$  & $75.0$  & $\textbf{53.7}$  & $20.6$  & $26.4$  & $\textbf{23.4}$  & $20.3$  & $21.3$  & $20.8$ \\
     
\end{tabular}
}
\caption{\revise{Zero-Shot Learning results from our \texttt{APN} model and other state-of-the-art on CUB, AWA2, and SUN datasets. All the models are trained with unsupervised class embedding, i.e. w2v.
\texttt{SJE}~\citep{akata2015evaluation}, \texttt{GEM-ZSL}~\citep{liu2021goal}, and our model \texttt{APN} are non-generative models, while \texttt{CADA-VAE}~\citep{schonfeld2019} is feature generation model. We measure top-1 accuracy~(\textbf{T1}) in ZSL, top-1 accuracy on seen/unseen~(\textbf{s/u}) classes and their harmonic mean~(\textbf{H}) in GZSL. }
}
\label{tab:ZSL}
\end{table*}

\begin{table*}[h]
\centering
\resizebox{\linewidth}{!}{%
    \begin{tabular}{c | x{0.8cm} x{0.8cm} x{0.8cm} x{0.8cm} x{0.8cm} | x{0.8cm} x{0.8cm} x{0.8cm} x{0.8cm} x{0.8cm}}
    % \toprule
        \multirow{2}{*}{\textbf{Method}} & \multicolumn{5}{c|}{\textbf{Few-Shot Learning}}  & \multicolumn{5}{c}{\textbf{Generalized Few-Shot Learning}}
        \\ 
        & \textbf{1} & \textbf{2} & \textbf{5} & \textbf{10} &\textbf{20} & \textbf{1} & \textbf{2} & \textbf{5} & \textbf{10} & \textbf{20}  \\ \hline
\texttt{softmax} &  49.3   &  64.5   &   76.7  &  81.0   &  84.2   &  50.0   &  60.8   &  73.5   &  79.0   &  80.7 \\
\texttt{Analogy}~\citep{hallucinate2017low} &  40.5   &   50.7  &    61.6 &   69.5  &  76.0   &  51.5   &  59.8   &  67.4   &  72.0   &  77.3  \\
\texttt{f-VAEGAN-D2-ind}~\citep{xian2019}  &  54.4   &   64.4   &    74.6  &   79.9   &   84.0   &   60.3   &   65.7   &   73.5   &    78.8  &    79.5   \\
\texttt{f-VAEGAN-D2}~\citep{xian2019}  &  60.1   &  70.0   &  79.0   &  81.5   &   84.5  &   66.3  &  72.6   &  78.6   &  81.5   &  83.2 \\
\texttt{APN(w2v)+f-VAEGAN-D2}  &  \textbf{61.7}   &  \textbf{70.9}   &  \textbf{79.5}   &   \textbf{83.4}  &   \textbf{85.5}  &  \textbf{67.5}   &   \textbf{73.8}  &  \textbf{79.6}  &  \textbf{82.5}   &  \textbf{84.3} \\
      
    % \bottomrule
    \end{tabular}}
    \caption{\revise{Few-Shot Learning results on ImageNet with increasing number of training samples per novel class (top-5 accuracy). Our \texttt{APN(w2v)} model is trained with w2v class embeddings. We apply our APN features to feature generation model~(i.e.  \texttt{APN(w2v)+} \texttt{f-VAEGAN-D2}).}}
    \label{tab:imagenet}
\end{table*}

\new{As shown in Table~\ref{tab:meta_cub}, in the N-way-K-shot scenario where we train \texttt{DPGN}~\citep{yang2020dpgn} with our attribute prototype network, we yield improvement over the baseline \texttt{DPGN} on CUB and achieve new state-of-the-art accuracy. We achieve $77.4\%$~(5way-1shot), compared to $68.4\%$~(\texttt{CloserLook}), $74.1\%$~(\texttt{RAP+ProtoNet}) and $75.7\%$~(\texttt{DPGN}). In the 5way-5shot setting, we also improve over \texttt{CloserLook}, \texttt{RAP+ProtoNet} and \texttt{DPGN} by $12.9\%$, $3.0\%$ and $0.7\%$ respectively. The results indicate that integrating attribute prototype network into the representation learning process helps the network to learn locality enhanced features and benefit the FSL performance.}

\subsection{Discussion}

\revise{We study the flexibility of our \texttt{APN} network by using only unsupervised embeddings, e.g. word2vec~(w2v)~\citep{mikolov2013distributed}. Similar to learning attributes, for w2v with 300 dim, we design 300 prototypes in the \texttt{ProtoMod}. Since there is no grouping for w2v, we discard the attribute decorrelation loss $\mathcal{L}_{\Scale[0.6]{AD}}$ when training the model.}

\revise{We compare the zero-shot learning results of our APN model with other state-of-the-art models in Table~\ref{tab:ZSL}. The performance of using w2v as the class embeddings could drop compared to attributes. But this issue exists in most zero-shot learning methods, as shown in Table~\ref{tab:ZSL}. In general, our method can be generalized to unsupervised class embeddings and still outperforms the baselines using w2v.
For instance, we outperform other non-generative models such as \texttt{SJE} and \texttt{GEM-ZSL} on three datasets, e.g. we achieve 59.6\% on AWA2, comparing with \texttt{SJE} with 53.7\% and \texttt{GEM-ZSL} with 50.2\%. Our model even outperform generative model \texttt{CADA-VAE} that synthesize image features on AWA2 and CUB dataset.}

\revise{We perform few-shot learning on ImageNet to study the generalization ability of our model on the large-scale dataset. We follow the data split in~\citet{xian2019} where the 1K ImageNet categories are randomly divided into 389 base classes and 611 novel classes. We use the 300-dim word2vec~\citep{mikolov2013distributed} embeddings as the class embedding for ImageNet since there is no attribute annotation. Our \texttt{APN} model is trained with train samples from the base classes. 
The results shown in Table~\ref{tab:imagenet} demonstrate that our \texttt{APN} model trained with only w2v learns better representations and further boosts the few-shot learning performance on the large-scale ImageNet dataset. Notably, we gain 1.6\%~(1-shot) on FSL and 1.2\%~(1-shot) on GFSL.}

\section{Conclusion}
\add{In this work, we develop a representation learning framework for zero-shot learning and few-shot learning, i.e. attribute prototype network (\texttt{APN}), to jointly learn global and local features.
Our model improves the locality of image representations by regressing attributes with local features and decorrelating 
% \bernt{the term `decorating' is not appropriate here - and i am not even sure what you want to say. Maybe `learning'?} 
prototypes with regularisation. }\new{We explicitly encourage the network to learn from informative attribute related image regions and discard noisy backgrounds by cropping the original image with attribute similarity maps.} \add{
We demonstrate consistent improvement over the state-of-the-art on three benchmarks. And our representations improve over finetuned ResNet representations when used in conjunction with feature generating models. We qualitatively verify that our network is able to localize attributes in images accurately. Two well-designed user studies indicate that our network can generate semantically consistent and accurate attribute attention maps. The part localization accuracy significantly outperforms a weakly supervised localization model designed for zero-shot learning. We further show that our model can be extended to the FSL scenario, and we consistently improve the classification accuracy in any-shot regimes on three datasets.}

%----------Acknowledgements--------------

\begin{acknowledgements}
This work has been partially funded by the ERC 853489 - DEXIM and by the DFG – EXC number 2064/1 – Project number 390727645.
\end{acknowledgements}

% \ifCLASSOPTIONcompsoc

% \clearpage

\bibliographystyle{spbasic}      % basic style, author-year citations
\bibliography{mybib}   % name your BibTeX data base
% \bibliography{mybib}
% \begin{thebibliography}{}
% %
% % and use \bibitem to create references. Consult the Instructions
% % for authors for reference list style.
% % Format for Journal Reference
% Author, Article title, Journal, Volume, page numbers (year)
% Author, Book title, page numbers. Publisher, place (year)
% \end{thebibliography}

\end{document}